\documentclass[5p, final, times]{elsarticle}

\usepackage{lineno,hyperref}
\usepackage{url}

\usepackage{textcomp}
\usepackage{xcolor}
\usepackage{amsmath,amssymb,amsfonts}
\usepackage{algorithmic}
\usepackage{graphicx}
\usepackage{subcaption}

\biboptions{sort&compress}

\begin{document}

\newtheorem{assumption}{Assumption}

\begin{frontmatter}

\title{Adaptive Least Mean $p^{th}$ Power Graph Neural Networks}

\author[1]{Yi Yan\corref{cor1}
}
\ead{pcr21@mails.tsinghua.edu.cn}

\author[1]{Changran Peng\corref{cor1}
}
\ead{y-yan20@mails.tsinghua.edu.cn}

\author[1]{Ercan E. Kuruoglu\corref{cor2}
}
\ead{kuruoglu@sz.tsinghua.edu.cn}

\address[1]{Tsinghua-Berkeley Shenzhen Institute, Shenzhen International Graduate Scool, Tsinghua University, Shenzhen, China.}
\cortext[cor1]{Equal contribution}
\cortext[cor2]{Corresponding author}

\begin{abstract}
In the presence of impulsive noise, and missing observations, accurate online prediction of time-varying graph signals poses a crucial challenge in numerous application domains. We propose the Adaptive Least Mean $p^{th}$ Power Graph Neural Networks (LMP-GNN), a universal framework combining adaptive filter and graph neural network for online graph signal estimation. LMP-GNN retains the advantage of adaptive filtering in handling noise and missing observations as well as the online update capability. The incorporated graph neural network within the LMP-GNN can train and update filter parameters online instead of predefined filter parameters in previous methods, outputting more accurate prediction results. The adaptive update scheme of the LMP-GNN follows the solution of a $l_p$-norm optimization, rooting to the minimum dispersion criterion, and yields robust estimation results for time-varying graph signals under impulsive noise. A special case of LMP-GNN named the Sign-GNN is also provided and analyzed, Experiment results on two real-world datasets of temperature graph and traffic graph under four different noise distributions prove the effectiveness and robustness of our proposed LMP-GNN.

\end{abstract}

\begin{keyword}
Graph neural networks, Graph signal processing, adaptive algorithms, impulsive noise, non-Gaussian noise
\end{keyword}

\end{frontmatter}

\section{Introduction}
Over recent years, there has been a notable increase in the development of innovative analysis techniques for interacting multivariate signals on irregular structures, giving rise to the emerging research field known as graph signal processing (GSP) \cite{Ortega_2018, dong2020graph}. 
In numerous applications related to sensor networks \cite{Spelta_2020_NLMS}, transportation \cite{yu2018_STGCN}, communication \cite{bib_LMS}, social interactions, or human organ systems \cite{brain_modeling}, the recorded data can be represented as signals defined across graphs, commonly referred to as graph signals \cite{Ortega_2018}. 
The objective of GSP is to expand traditional signal processing tools to the graph domain and conduct analysis of graph signals defined on an irregular discrete domain represented on graph structures. 
In GSP, the most important tool is graph spectral analysis, which is established based on the Graph Fourier Transform (GFT). 
The GFT is the graph analogy of the classical Fourier Transform; the projection of the graph signal onto the eigenvectors of the graph Laplacian matrix defines the graph spectrum, which helps us to view the graph signal via a spectral point of view \cite{Ortega_2018}.

In most real-world scenarios, data is time-varying, which leads to the task of how to better bridge the representation gap between the spatial domain and the temporal domain. 
GSP techniques have also been applied to the task of prediction of irregularly structured and time-varying data in many application fields, including wind speed prediction \cite{yan_2023_diffusion}, brain connectivity analysis \cite{zhao2023sequential}, and traffic flow estimation \cite{yu2018_STGCN}. 
One possible solution is to integrate time series analysis techniques with GSP methods.
For instance, GSP can be integrated with the Vector Autoregressive model \cite{Mei_GVAR_2017}, the Vector Autoregressive–Moving-Average model \cite{Isufi_GARMA_2017}, and the GARCH model \cite{Hong_GGARCH_2023}.
However, collected data often suffers from the problem of missing observation and contamination by noise, which makes the learning of the model parameters in time-series models difficult, not to mention the high model complexity. 
In classical signal processing, there is another line of work known as the adaptive filter, which provides an effective and minimalist model for the prediction of time-varying signals.
Additionally, the adaptive filters can make predictions in real-time as they receive new data observations, which is also known as online estimation \cite{bib_classical_adaptive_filter}.
Combining GSP and adaptive filtering, the adaptive graph Least Mean Squares (GLMS) algorithm was first proposed. 
The algorithm employed a fixed, predefined bandlimited filter acquired from the GFT to update, utilizing an LMS-based update term \cite{bib_LMS}. 
Since then, various improved versions of GLMS have been proposed.
The Normalized GLMS (GNLMS) algorithm speeds up the convergence behavior of the GLMS \cite{Spelta_2020_NLMS}.
The graph (unnormalized and normalized) least means pth algorithm was proposed to aim at the stable operation under Symmetric Alpha Stable (S$\alpha$S) noise \cite{nguyen2020_LMP}.
The Graph-Sign algorithm has fast operation time and has robust performance under various non-Gaussian impulsive noise \cite{yan_2022_sign}. 

\begin{figure}[h]
     \centering
     \begin{subfigure}{\linewidth}
         \centering
         \includegraphics[width=\textwidth]{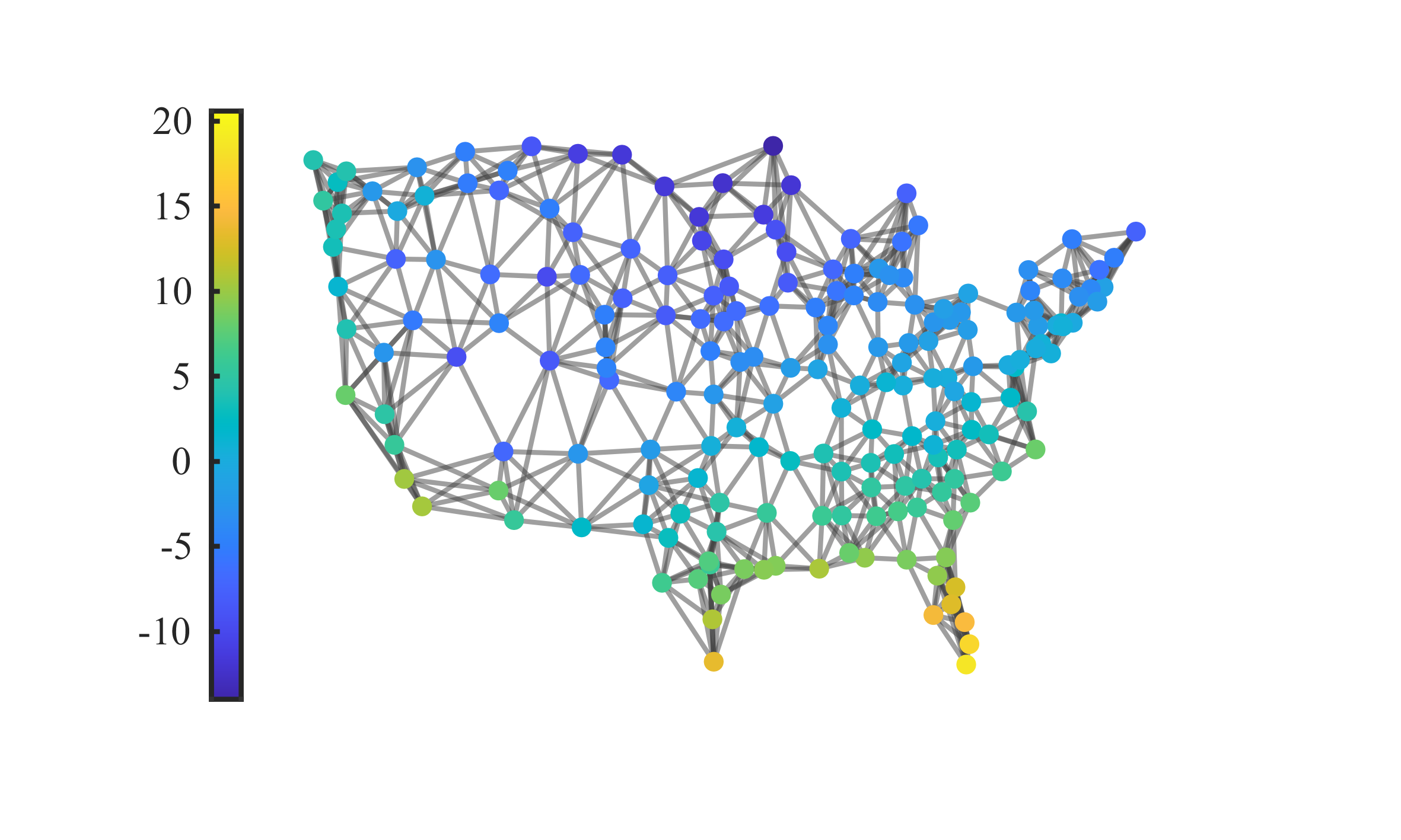}
     \end{subfigure}
     
     \vspace{-30 pt}
     
     \begin{subfigure}{\linewidth}
         \centering
         \includegraphics[width=\textwidth]{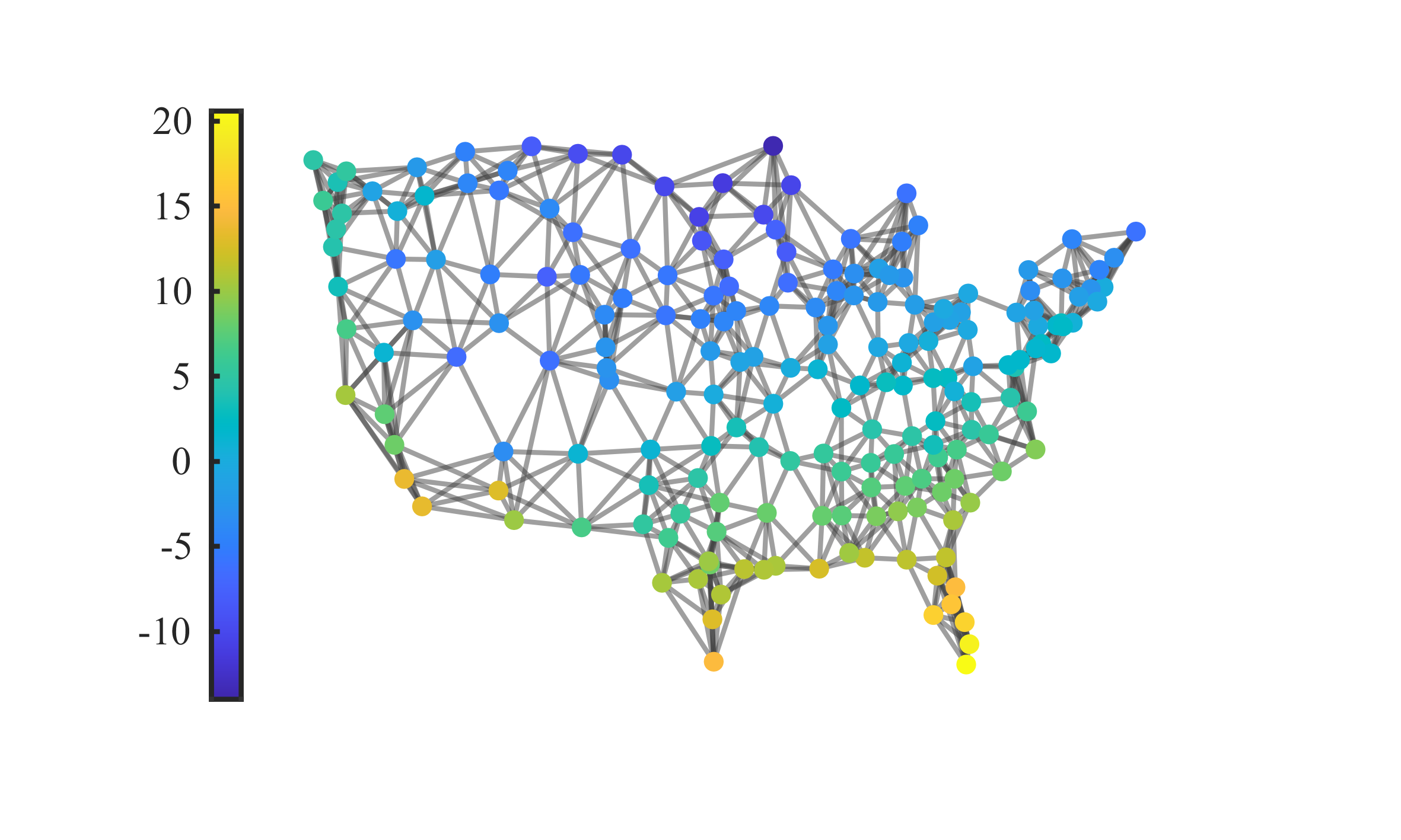}
     \end{subfigure}
     
      \vspace{-30 pt}

      \begin{subfigure}{\linewidth}
         \centering
         \includegraphics[width=\textwidth]{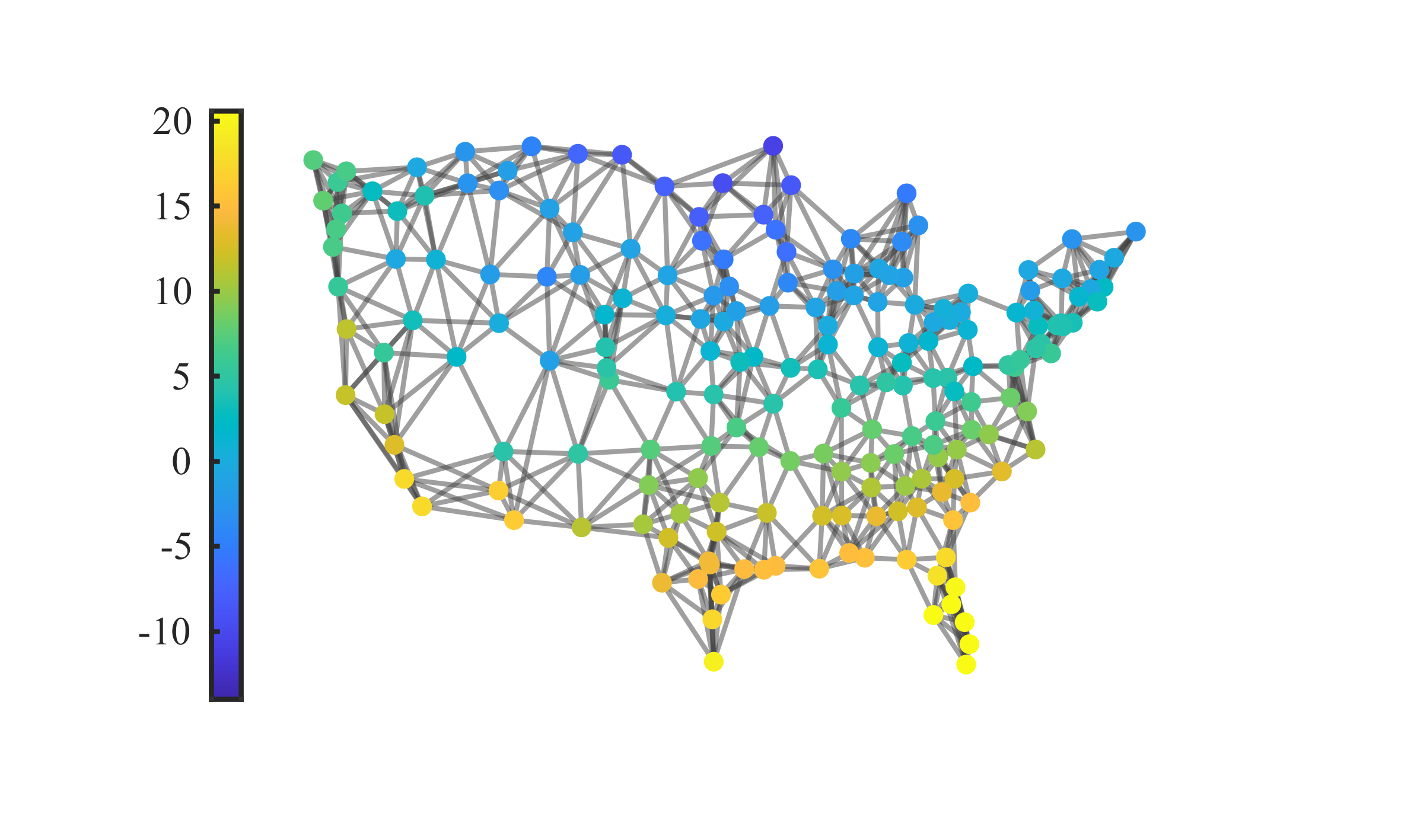}
     \end{subfigure}
     
     \vspace{-20 pt}
     
        \caption{An illustration of the temperature graph with time-varying temperature recordings at each weather station, displayed at three different time instances.}
        \label{fig_ground_truth}
\end{figure}

Although previously mentioned GSP methods have achieved relatively good results, the results rely heavily on the accurate design of a fixed predefined filter, which in reality, usually requires human prior knowledge varying in different tasks under different situations.
Graph Neural Network naturally extends GSP techniques from both spatial and spectral domains to machine learning tasks like link prediction and node classification \cite{bib_GCN, bruna2013_spectral_GCN}. 
However, most tasks that GNNs have been proved successful are time-invariant, which means the data does not have time-varying features. 
The Spatio-Temporal GCN (STGCN) leverages the combination of GCN and gated CNN to process spatial features and temporal features, thus performing better \cite{yu2018_STGCN}. Several variants of STGCN further improve the prediction performance of STGCN by leveraging synchronous modeling mechanism \cite{song2020spatial} and generating temporal graph \cite{li2021spatial}.
However, the STGCN does not take into account the problem of data regarding missing observations and the existence of noise. Moreover, the STGCN requires a huge amount of training data and is an offline method in which all training is finished before deployment.

The Gaussian assumption is usually the first choice of noise models and can be found in various classical and GSP adaptive filtering algorithms \cite{bib_classical_adaptive_filter, bib_LMS, Spelta_2020_NLMS}. 
However, an oversimplification of directing to the Gaussian model regardless of the application scenario is superficial.
This is not to mention that in certain algorithm designs, data is assumed to be clean, and noise is a factor that was neglected. 
For communication applications such as underwater communications and powerline communications, the Cauchy-Gaussian mixture model and the Student's t-distribution can be adopted as noise models \cite{bib_underwater, bib_student_t_noise, b17_PLC}.
In image processing, $\alpha$-stable noise has been shown to better model the behavior of noise compared to the Gaussian distribution \cite{idan2010cauchy, herranz2004alpha}.
The presence of impulsive noise will cause instabilities for algorithms with Gaussian noise assumptions and algorithms with clean data assumptions due to the outliers introduced by the heavy-tailed behaviors of the impulsive noise \cite{bib_MD_LMAD, nguyen2020_LMP}. 
The presence of impulsive noise in real-life applications creates a demand for algorithms that are stable and robust.

In this paper, we propose the Adaptive Least Mean $p^{th}$ Power Graph Neural Networks (LMP-GNN), which performs robust online prediction of time-varying graph signals with missing observations and under noise pollution. 
Instead of fixed predefined filters seen in previous work, LMP-GNN utilizes the filter learned by a neural network as a time-varying bandlimited filter, therefore performing better prediction. 
On one hand,  LMP-GNN is born with the characteristics of graph adaptive filters to update prediction according to the error between prediction and observation in an online manner.
On the other hand, LMP-GNN inherits the ability of neural networks to train and learn the parameters from the signals themselves. 
Experiments on real-world temperature data and traffic data demonstrate that LMP-GNN indeed combines the advantages of both graph adaptive filter methods and graph neural network methods by outperforming both.

The main contributions of our work can be summarized as follows:
\begin{itemize}
    \item The LMP-GNN is a universal framework combining adaptive filter and graph neural network for the task of online estimation of time-varying graph signal. 
\end{itemize}

\begin{itemize}
    \item A special case of LMP-GNN is also being proposed and analyzed: the Sign Graph Neural Networks (Sign-GNN). 
\end{itemize}

\begin{itemize}
    \item Tested under four different noise distributions, Student's t-distribution, $\alpha$-stable distribution, Laplace distribution, and Cauchy distribution, the LMP-GNN demonstrates robustness under various noises.
\end{itemize}

The rest of the paper is organized as follows. Section~\ref{sec_Background} goes over the preliminary knowledge of GSP and several probability distributions as noise models. In Section~\ref{sec_method}, we will give a detailed derivation of our proposed LMP-GNN and its special case Sign-GNN. All experimental results are shown and discussed in Section~\ref{sec_experiments}. We form our conclusion in Section~\ref{sec_conclusion}.

\section{Background}
\label{sec_Background}

\subsection{Impulsive Distributions}

\begin{figure*}[h]
    \centering
    \begin{subfigure}{0.495\textwidth}
        \centering
        \includegraphics[width=\textwidth]{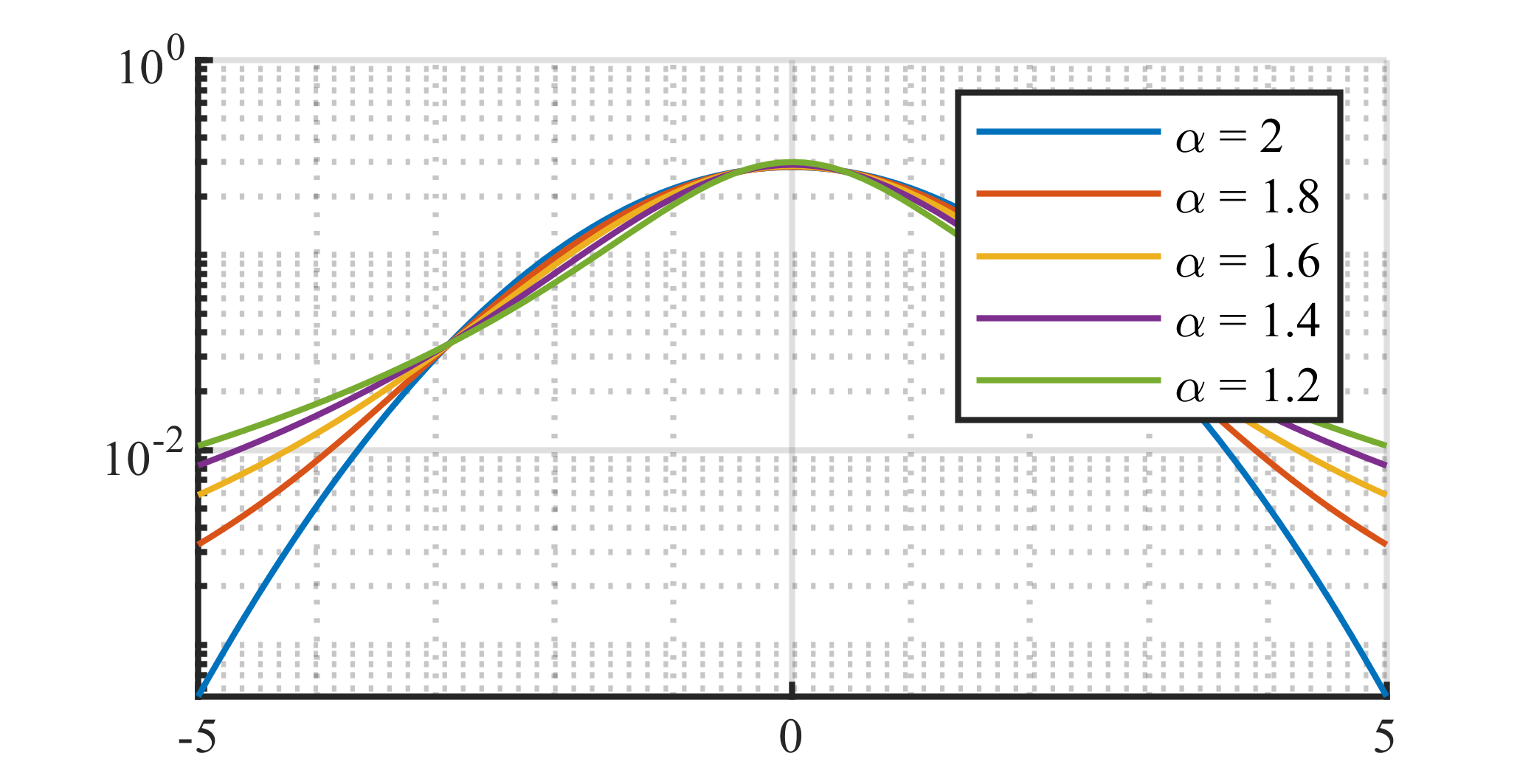}
        \caption{The S$\alpha$S distribution with $\gamma$ = 1 and various $\alpha$ values.}
        \label{PDF_SAS}
    \end{subfigure}
        \begin{subfigure}{0.495\textwidth}
        \centering
        \includegraphics[width=\textwidth]{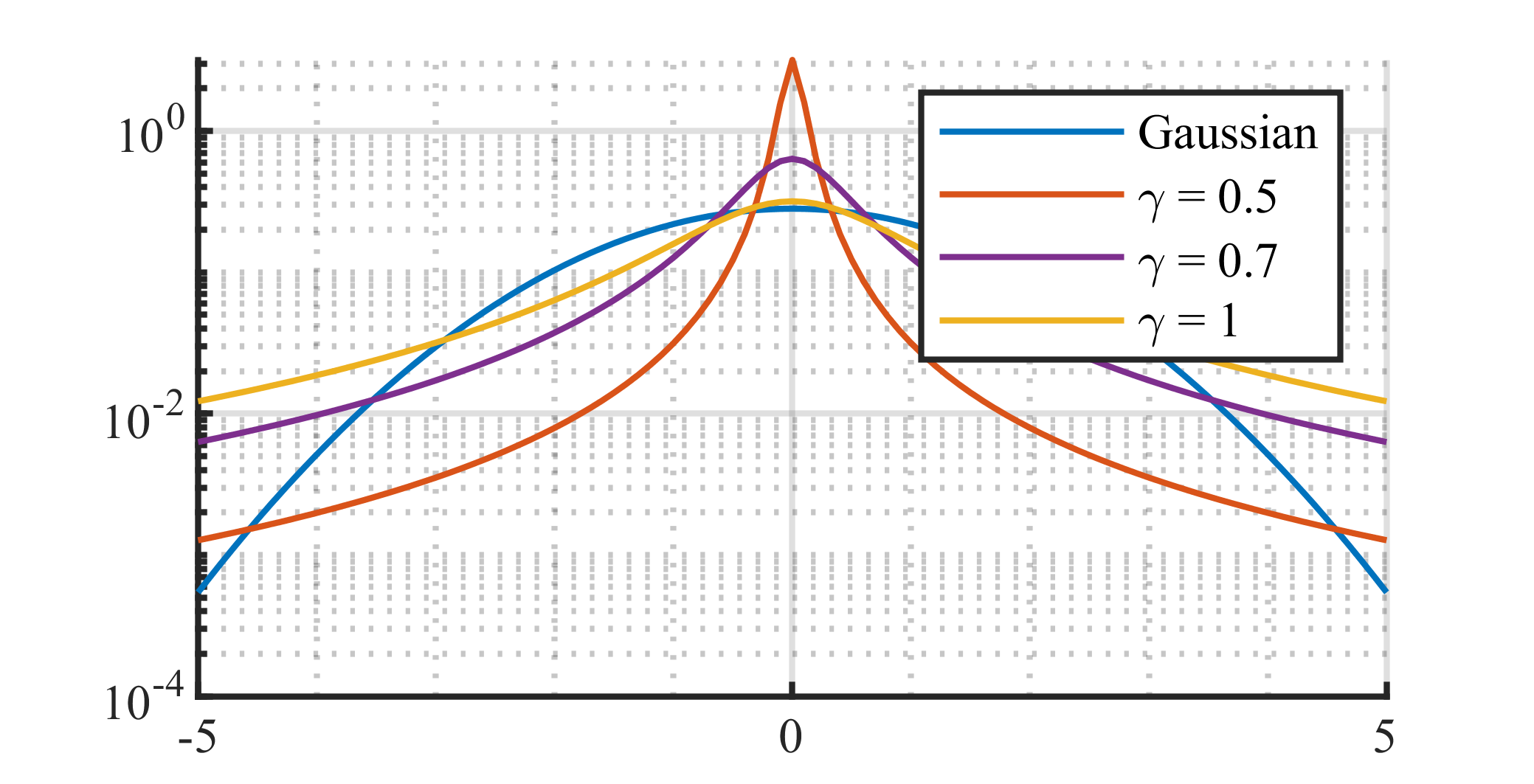}
        \caption{The Cauchy distribution with various $\gamma$ values and the Gaussian distribution.}
        \label{PDF_Cauchy}
    \end{subfigure}
    \vskip 0\baselineskip
    \begin{subfigure}{0.495\textwidth}
        \centering
        \includegraphics[width=\textwidth]{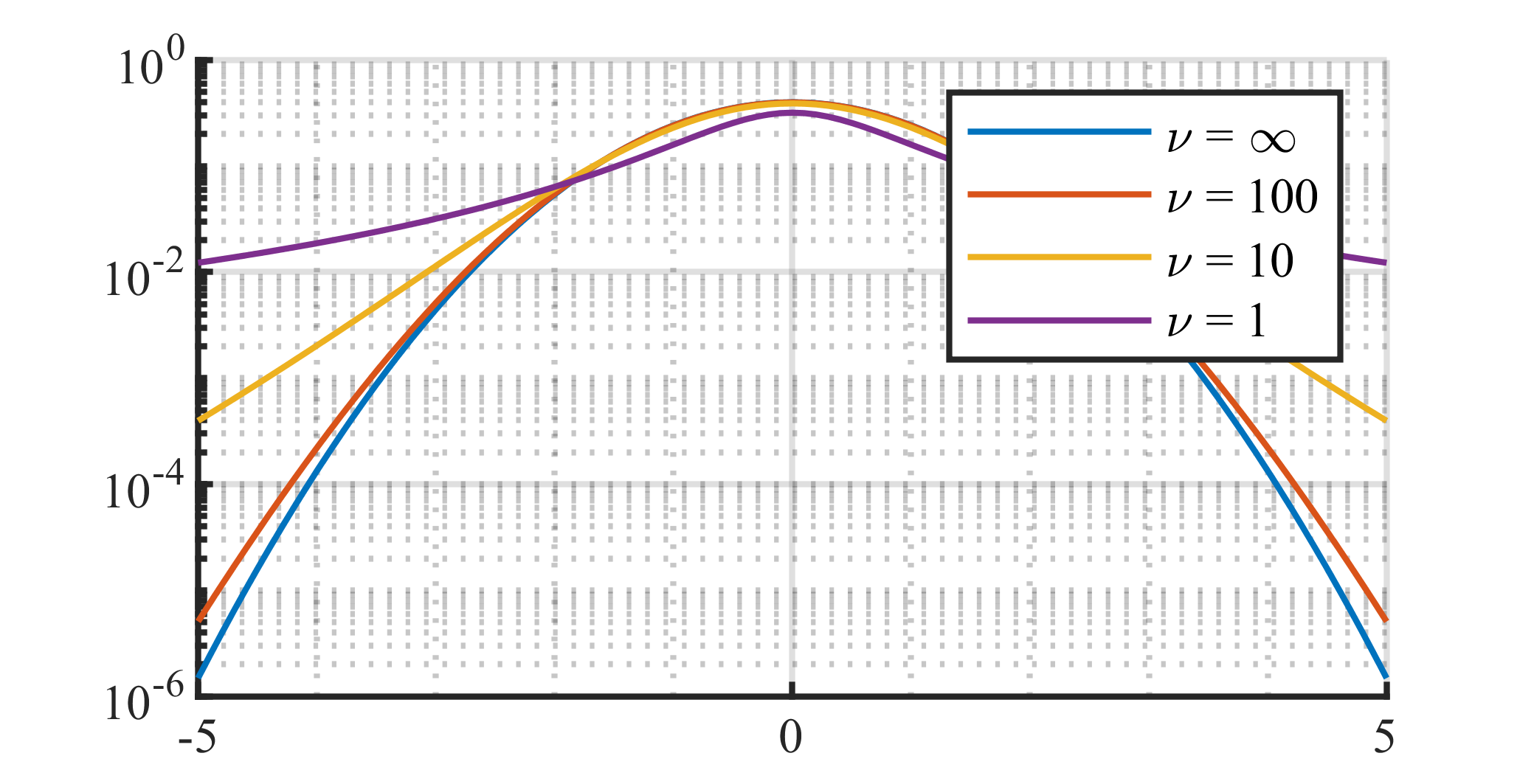}
        \caption{The Student's t-distribution with various $\nu$ values.}
        \label{PDF_ST}
    \end{subfigure}
    \begin{subfigure}{0.495\textwidth}
        \centering
        \includegraphics[width=\textwidth]{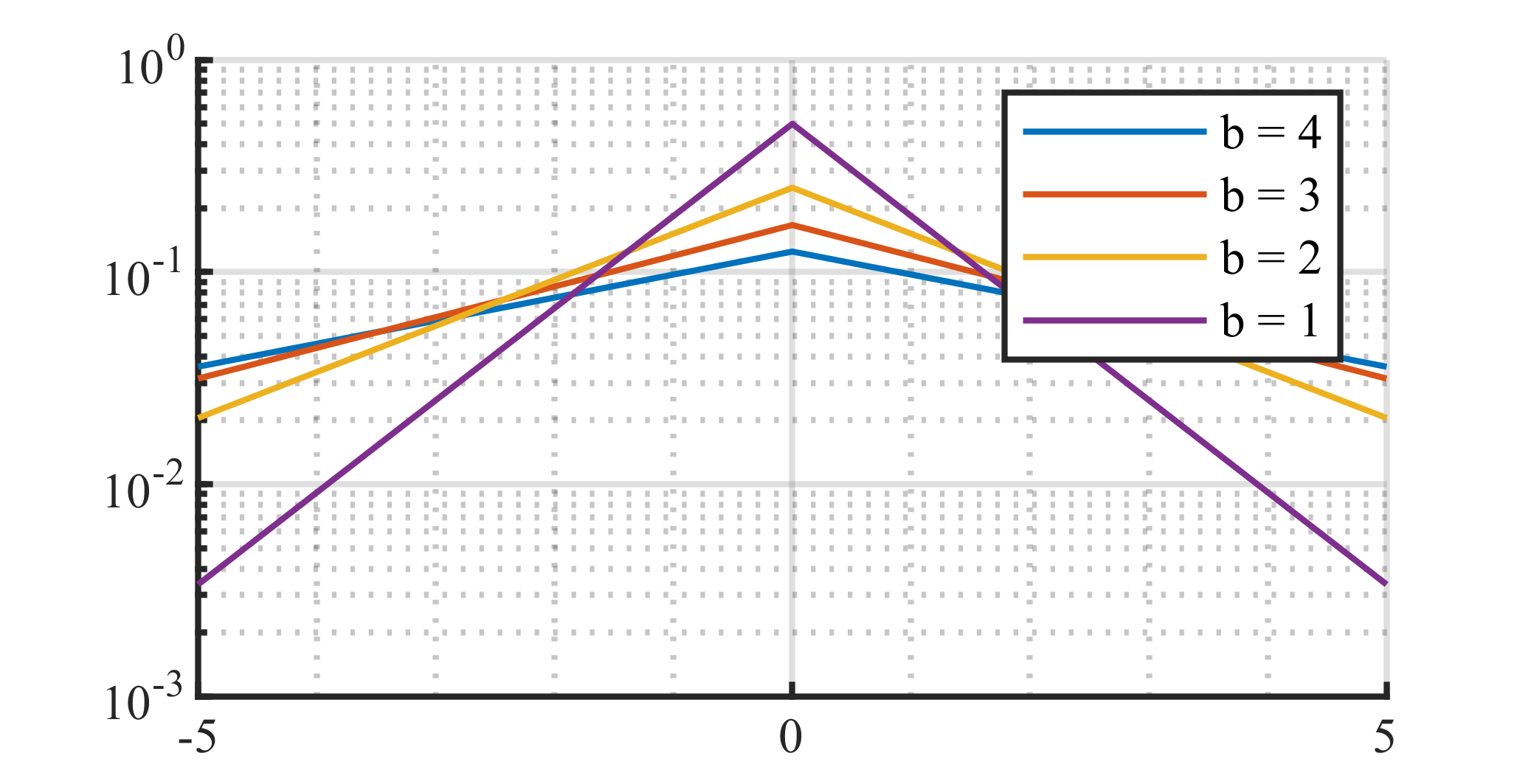}
        \caption{The Laplace distribution with various $b$ values.}
        \label{PDF_LP}
    \end{subfigure}    
    \caption{The probability density functions of various non-Gaussian distributions all with zero-mean/median (log scale).}
    \label{PDF}
\end{figure*}

We will be considering five cases of noise distributions in this paper: the Gaussian distribution, the S$\alpha$S distribution, the Cauchy distribution, the Student's t-distribution, and the Laplace distribution.
In later sections, the noise distributions will be blended into the graph signals as additive noise.

Let us first begin with the Gaussian distribution, which is also known as the normal distribution. 
The probability density functions of the Gaussian distribution is 
\begin{equation}
    f(x) = \frac{1}{\sigma \sqrt{2\pi}}e^{-\frac{1}{2}\left(\frac{(x-\mu_\alpha)}{(\sigma)}\right)^2},
    \label{eq_gaussian_pdf}
\end{equation}
where $\mu$ is the mean and the median and $\sigma$ is the standard deviation. 
One important property of the Gaussian distribution is that it obeys the central limit theorem.
It is also worth mentioning that the Gaussian distribution is a special case of the S$\alpha$S distribution. 

The S$\alpha$S distribution is controlled by three parameters: the characteristic exponent parameter $\alpha$, the location parameter $\mu_\alpha$, and the scale parameter $\gamma$.  
One important property of the S$\alpha$S distribution is that the S$\alpha$S distribution has no analytic probability density functions for $\alpha$ values other than $\alpha = 2$ or $\alpha = 1$.
We will be relying upon the characteristic function in place of the probability density functions:
\begin{equation}
    \boldsymbol{\phi}(x)=\exp\left\{j\mu x-\gamma|x|^\alpha\right\}.\label{SaS}
\end{equation}
The impulsiveness of the S$\alpha$S distribution is dictated by $\alpha$, ranging from $0< \alpha \leq 2$, the impulsiveness decreases as $\alpha$ increases.
The mean and median of S$\alpha$S is $\mu$ only when $\alpha>1$. 
In the case where $\alpha<1$, $\mu_\alpha$ is only the median, the mean is undefined. 
As for the variance of S$\alpha$S, it is undefined unless for $\alpha$ = 2. 
Similar to the Gaussian distribution, the S$\alpha$S distribution also satisfies the central limit theorem \cite{kuruoglu1997new}.
We can model the Gaussian distribution by setting  $\alpha = 2$ to the S$\alpha$S distribution.

The Student's t-distribution is an impulsive distribution with heavy-tailed behavior for some parameter settings of the degrees of freedom $\nu$, with probability density functions of Student's t-distribution defined as
\begin{equation}
    \boldsymbol{f}(t)=\frac{\Gamma\left(\frac{\nu+1}{2}\right)}{\sqrt{\nu\pi}\Gamma\left(\frac{\nu}{2}\right)}\left(1+\frac{t^2}{\nu}\right)^{\frac{-\nu+1}{2}}, \label{student_t}
\end{equation}
where $\Gamma$ is the gamma function.
The purpose of the Student's t-distribution is to estimate the mean of Gaussian distribution $\mu_\alpha$, under a small sample size and an unknown variance. 
The parameter $\nu$ in the Student's t-distribution is the distribution of the location of the sample mean of the data relative to the true mean of the underlying Gaussian distribution that is divided by the data sample standard deviation. 
The Student's t-distribution will have infinite variance for parameter settings $1<\nu\leqslant2$, and will have undefined variance for parameter settings  $\nu\leqslant1$. 
However, when the parameter $\nu = \infty$, the Student's t-distribution is the Gaussian distribution.  

The Cauchy distribution, with its behavior also being impulsive and heavy-tailed, is the special case of both the S$\alpha$S distribution when $\alpha = 1$ and the Student's t-distribution when $\nu=1$. 
The probability density functions of the Cauchy distribution is
\begin{equation}
    \boldsymbol{f}(t,\gamma)=\frac{1}{\pi\gamma\left[1+\left(\frac{t-\mu_\alpha}{\gamma}\right)^2\right]}.\label{cauchy}
\end{equation}

Finally, the Laplace distribution, or the double exponential distribution, is a special case of generalized Gaussian distribution and is characterized by the location parameter $\mu$ and the scale parameter $b$. 
The probability density functions of the Laplace distribution can be expressed as
\begin{equation}
    \boldsymbol{f}(t,\mu,b)=\frac{1}{2b}\exp\left(-\frac{|t-\mu|}{b}\right).\label{laplace}
\end{equation}

An illustration of the Probability Density Functions of the above five distributions is given in Fig.~\ref{PDF}. 
By inspecting Fig.~\ref{PDF}, heavy-tailed behavior can indeed be observed for certain parameter settings of the non-Gaussian distributions.
When these distributions are used as noise models, they will generate outliers in the observation.


\subsection{GSP Preliminaries}

A time-varying graph signal $\boldsymbol{x}[t]$ is a multivariate function value with $N$ variables that change over time defined on the nodes of a graph $\mathcal{G}$. 
This means that there will be $N$ nodes on the graph $\mathcal{G}$ which forms a node set $\mathcal{N}$. 
The nodes in a graph are connected through the edges of the graph, and in this paper the edges are undirected.
A time-varying graph signal of hourly temperature recorded at $N = 197$ weather stations on a temperature graph is shown in Figure~\ref{fig_ground_truth}. 

To represent the graph structure or the graph topology, we define the graph adjacency matrix $\mathbf{A}$.
In $\mathbf{A}$, the $ij^{th}$ entry of $\mathbf{A}$ is the edge weight when there presents an edge between the $i^{th}$ node and $j^{th}$ node in $\mathcal{G}$.
In the case of an unweighted graph, we denote the edge weights as 1 for each edge appearance. 
The rest of the values in $\mathbf{A}$ are zeros.
The node degree matrix $\boldsymbol{D}$ is a diagonal matrix that records the node degree, i.e. the sum of the edge weights connecting to each of the nodes in $\mathcal{G}$.
For an unweighted and undirected graph, the degree of each node will be the number of edges each node has.

In order to conduct spectral domain operations on graphs using GSP, we need to define the GSP analogy of the Fourier Transform known as the GFT.
To begin the derivations of GFT, we begin with the difference between $\mathbf{D}$ and $\mathbf{A}$, which leads us to the graph Laplacian matrix $\mathbf{L}$ to be $\mathbf{L=D-A}$. 
Afterwards, the GFT is achieved by carrying out the eigenvector decomposition $\mathbf{L=U\Lambda U^{T}}$.
The matrix $\mathbf{\Lambda}$ is the diagonal matrix of eigenvalues recorded in increasing order $\boldsymbol{\lambda} = [\lambda_1, ... ,\lambda_N]^{T}$ to give a frequency analogy in GSP.
The corresponding eigenvector pair of each eigenvalue is in $\mathbf{U}$.
Since the GFT is the GSP analogy of the Fourier Transform, the forward GFT transforms a graph signal $\boldsymbol{x}$ from the spatial domain to the spectral domain: $\boldsymbol{s}=\mathbf{U}^{T}\boldsymbol{x}$.
Accordingly, the inverse GFT transforms $\boldsymbol{s}$ from the spectral domain back to the spatial domain: $\boldsymbol{x}=\mathbf{U}\boldsymbol{s}$.
Similar to the convolution property found in classical signal processing where convolution becomes multiplication in the spectral domain, we can apply a graph filter $\Sigma$ in the spectral domain to process graph signals: $\boldsymbol{s}'=\Sigma\boldsymbol{s}$.
Going along this direction, the detailed and complete GSP procedure to process a graph signal $\boldsymbol{x}$ using the filter $\mathbf{\Sigma}$ is 
\begin{equation}
    \boldsymbol{x}' = \mathbf{U}\mathbf{\Sigma}\mathbf{U}^{T}\boldsymbol{x}.
    \label{eq_GSP_filter}
\end{equation}

For the purpose of enforcing sparsity or modeling missingness in the graph signal, we can rely on the sampling operation. 
In other words, the sampling set $\mathcal{S}$ is a subset of $\mathcal{N}$ that records the nodes that are not missing.
Practically, the sampling matrix operation on the graph signals $\boldsymbol{x}$ can be achieved by a (diagonal) masking matrix $\mathbf{D}_\mathcal{S}$, with the diagonals defined as $\mathbf{D}_{\mathcal{S}_{ii}} = 1 \forall \, v_i \in \mathcal{S}$ \cite{bib_sampling}.

\section{Methodology}
\label{sec_method}

\subsection{Spectral Graph Neural Networks}
The LMP-GNN is a neural network algorithm closely related to the GCN.
Recalling the core of the GCN is the graph convolution achieved by an aggregation of graph features. 
In the GSP sense, the graph features are the graph signals, and graph convolutions in GCN bear the same formulation as the graph filtering in \eqref{eq_GSP_filter}:
\begin{equation}
    \text{agg}(\boldsymbol{x}) = \sum_{p=0}^{P} \theta_p \mathbf{L}^p \boldsymbol{x}\approx \mathbf{U} \sum_{f=1}^{F}\mathbf{\Theta}_f \mathbf{U}^T \boldsymbol{x} ,  
    \label{eq_agg}
\end{equation}
where $\boldsymbol{x}$ denotes graph signal, $\mathbf{\Theta}$ and $\theta$ are trainable parameters. 
Note that the expression on the right side of the approximation $\mathbf{U} \sum_{f=1}^{F}\mathbf{\Theta}_f \mathbf{U}^T \boldsymbol{x}$ is a spectral formulation of the GCN \cite{bruna2013_spectral_GCN}, while the expression on the left side of the approximation $\sum_{p=0}^{P} \theta_p \mathbf{L}^p \boldsymbol{x}$ is a spatial GCN \cite{kipf2016semi, defferrard2016_cheb}.
In the spectral GCN, each trainable parameter $\mathbf{\Theta}_f$ is a graph spectral filter similar to the GSP.
Take a closer look at the expression $\sum_{f=1}^{F}\mathbf{\Theta}_f$, we see that it is equivalent to the filter $\mathbf{\Sigma}$ in \eqref{eq_GSP_filter} if we linearly combine all $F$ filters by setting \begin{equation}
    \sum_{f=1}^{F}\mathbf{\Theta}_f = \mathbf{\Sigma},
    \label{eq_gcn_as_gsp}
\end{equation} giving the possibility of designing and analyzing GCNs from the spectral perspective and GSP techniques. 

Now, if we feed the aggregation in \eqref{eq_agg} into an activation function $\sigma()$, we can formulate a single layer of spectral GCN with $f$ filters as
\begin{equation}
    \boldsymbol{x}_{l+1}  = \sigma\left(\mathbf{U}\sum_{f=1}^{F}\mathbf{\Theta}_{f, l}\mathbf{U}^T\boldsymbol{x}_{l}\right),
    \label{eq_GNN}
\end{equation}
where $l$ denotes the layer index \cite{bruna2013_spectral_GCN}. We will denote the final layer as the $L^{th}$
layer.
If understanding the spectral GCN from a GSP point of view, the goal of each of the spectral GCN layers is to train a convolution by tuning trainable filters $ \mathbf{\Sigma} = \sum_{f=1}^{F}\mathbf{\Theta}_f$ in the spectral domain based on an optimized update scheme defined by a cost function.
If understanding the GCN from a deep learning perspective, the goal of each of the spectral GCN layers is to train the parameters $\sum^{F}_{f = 1}\mathbf{\Theta}_{f}$ using backpropagation, which will lead a $L$ layer GCN to train a sequence of filters that hopefully recovers the ground truth graph signal $\boldsymbol{x}_g$ by $\boldsymbol{x}_g = \boldsymbol{x}_L(...\boldsymbol{x}_1(\boldsymbol{x}_0))$. 

While GCN has been proven successful in numerous learning tasks, it assumes the cleanness of the training data and it was only proposed for the graph features that are time-invariant.
In the next subsections, we will briefly go over the processing of time-varying graph signals and how to process them using graph adaptive filters, then provide a detailed derivation that leads to our proposed LMP-GNN algorithm.

\subsection{Adaptive Filtering}
\begin{figure}[h]
     \centering
     \begin{subfigure}{\linewidth}
         \centering
         \includegraphics[width=\textwidth]{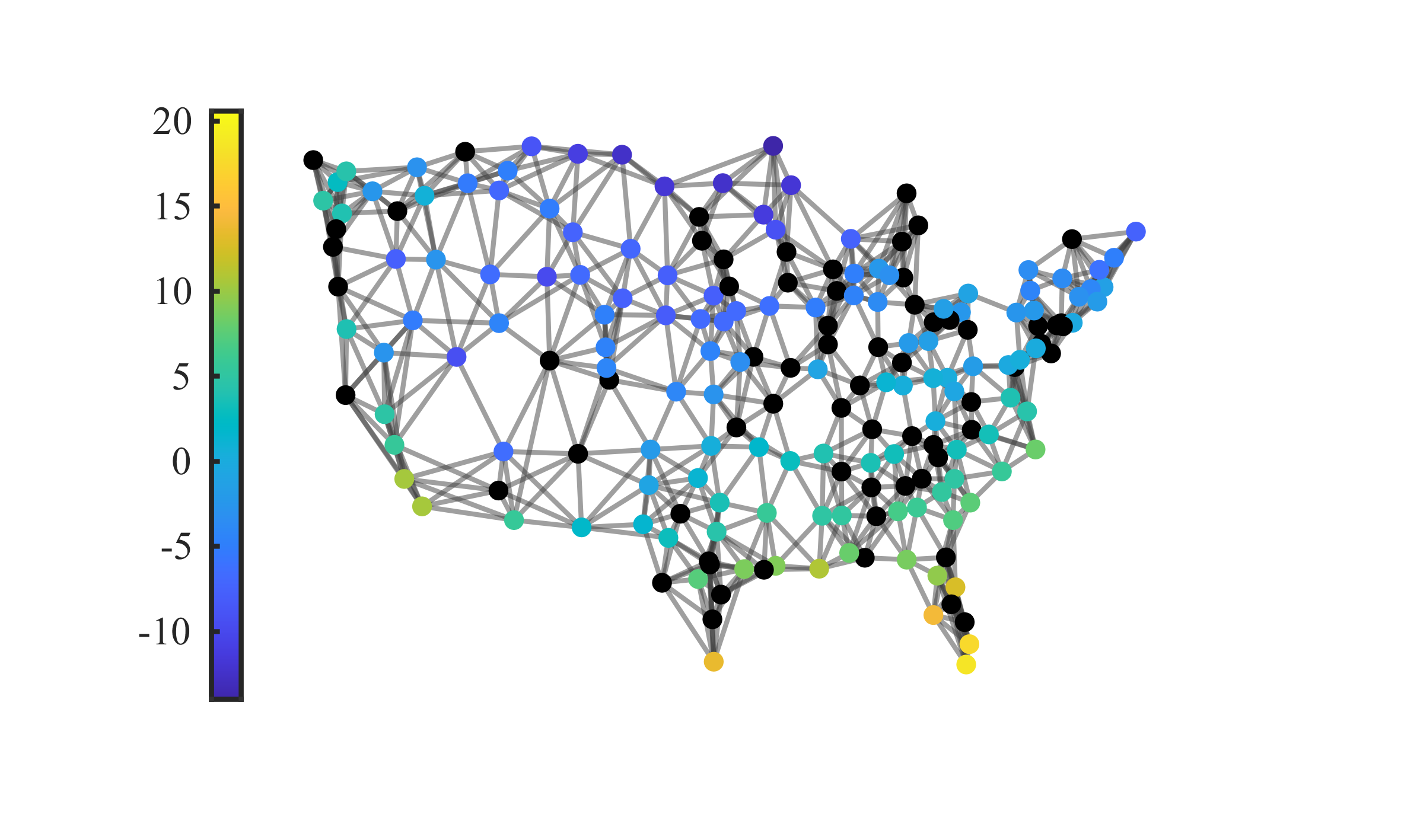}
     \end{subfigure}
     
     \vspace{-30 pt}
     
     \begin{subfigure}{\linewidth}
         \centering
         \includegraphics[width=\textwidth]{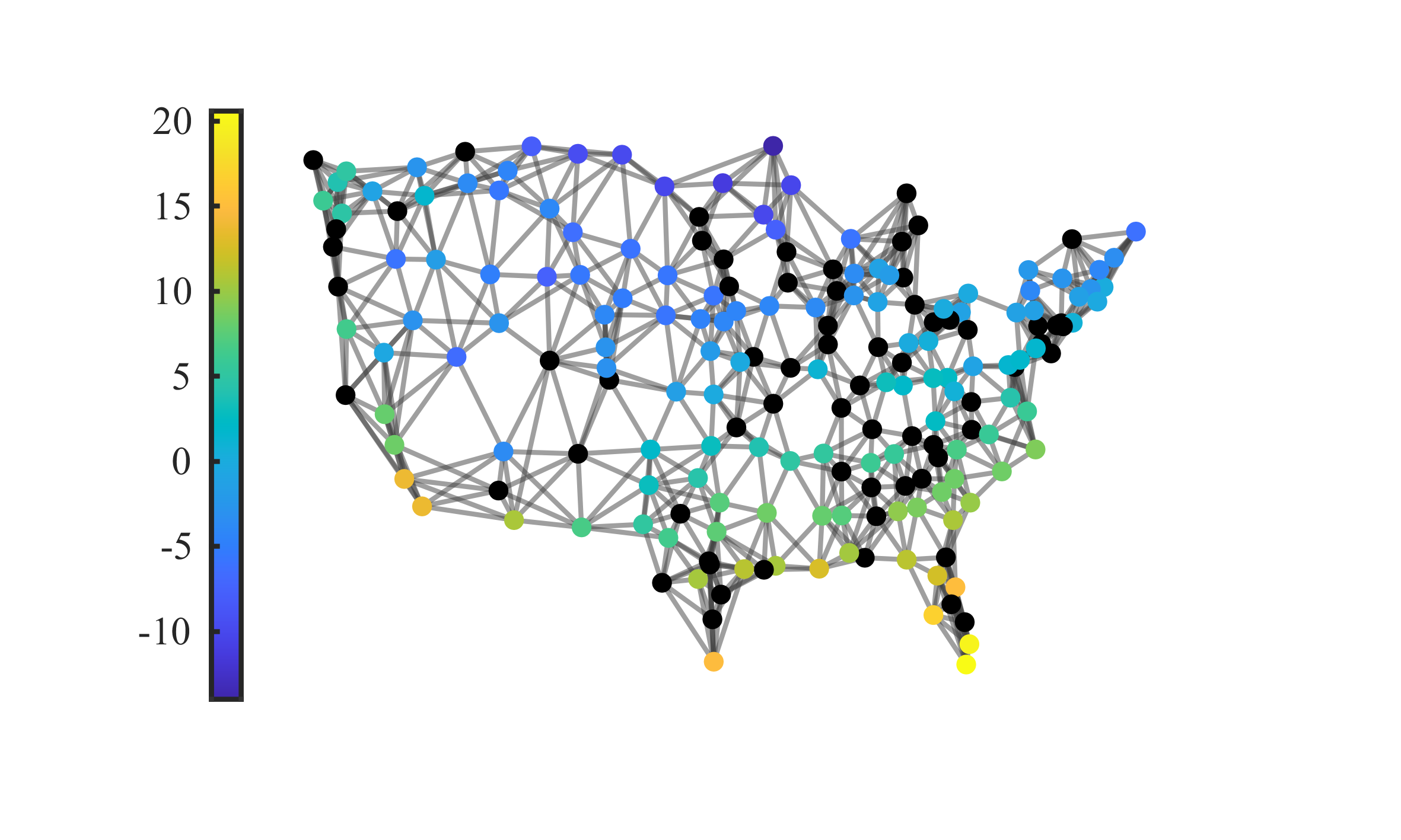}
     \end{subfigure}
          
     \vspace{-30 pt}
     
      \begin{subfigure}{\linewidth}
         \centering
         \includegraphics[width=\textwidth]{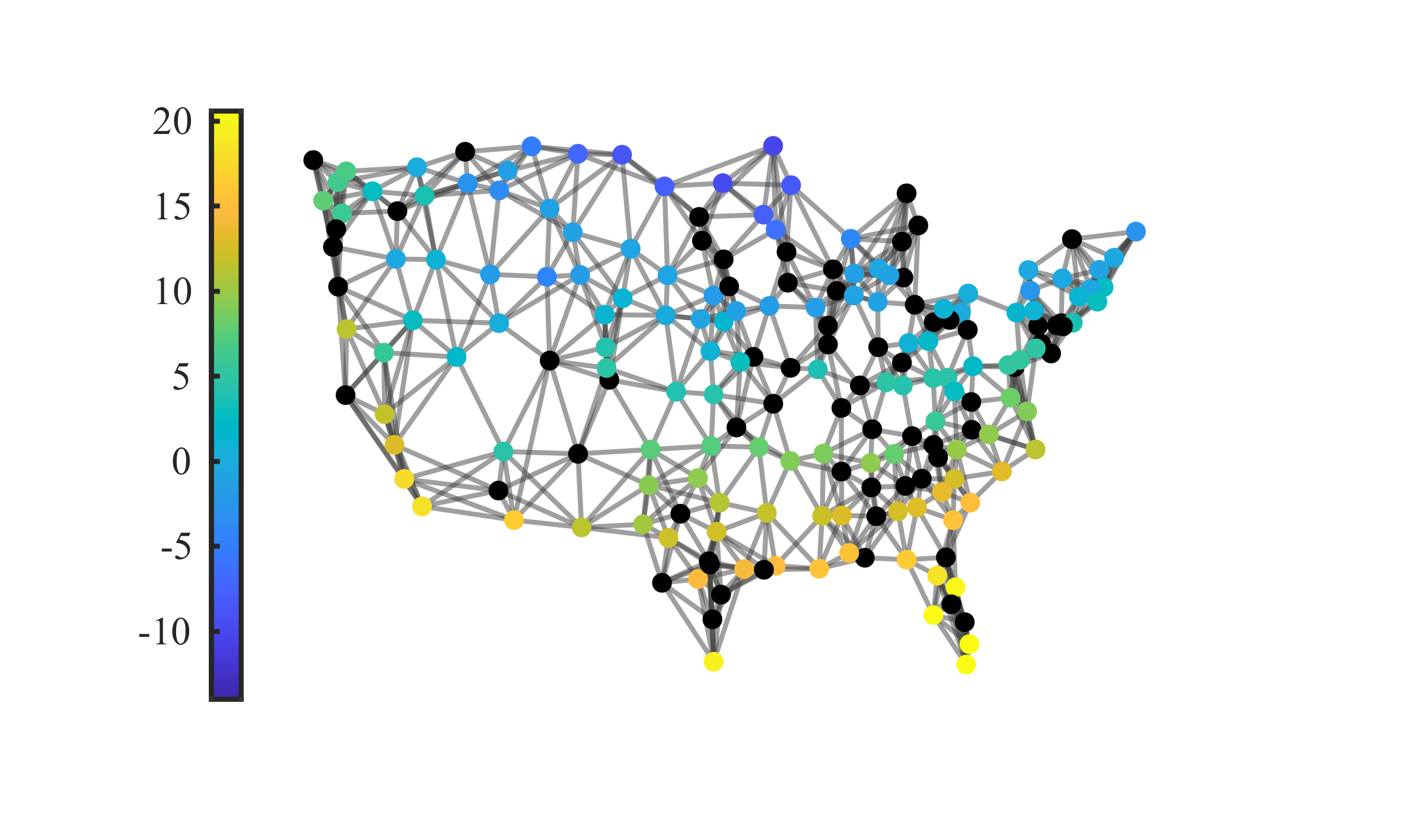}
     \end{subfigure}
          
     \vspace{-20 pt}
     
        \caption{A temperature graph with missing values and S$\alpha$S noise ($\alpha = 1.5$, $\gamma = 0.1$) on the time-varying temperature recordings at each weather station. The three displayed time instances are the same as in Figure~\ref{fig_ground_truth}.}
        \label{fig_missing}
\end{figure}

In graph signal estimation tasks, intuitively, what a graph adaptive filter does is try to predict the time-varying graph signal of the next time step given an observation of the current time step. 
The observation is assumed to contain noise and missingness.
Let $\boldsymbol{y}[t]$ denote such observation at time step $t$, with $\mathbf{D}_{\mathcal{S}}$ being the sampling mask matrix to represent missingness.
Thus, the observation model of the time-varying graph signal is 
\begin{equation}
    \boldsymbol{y}[t]=\mathbf{D}_{\mathcal{S}}(\boldsymbol{x}_g+\mathbf{w}[t]).
    \label{eq_observation}
\end{equation}
An illustration of a time-varying graph signal that has missing values and S$\alpha$S noise using the model in \eqref{eq_observation} is shown in Figure~\ref{fig_missing}. 

In order to complete the task of error minimization, the graph  adaptive filters will attempt to minimize the error between the current observation $\hat{\boldsymbol{x}}[t]$ and $\boldsymbol{y}[t]$:
\begin{equation}
\begin{aligned}
\min_{\hat{\boldsymbol{x}}[t]} \quad &  J(\hat{\boldsymbol{x}}[t]) =  C\left(\mathbf{D}_{\mathcal{S}}(\boldsymbol{y}[t]-\mathbf{U\Sigma U}^{T}\hat{\boldsymbol{x}}[t])\right),\\
\textrm{s.t.} \quad & \mathbf{U\Sigma U}^{T}\hat{\boldsymbol{x}}[t] = \hat{\boldsymbol{x}}[t],\\
\end{aligned}
\label{eq_optimization}
\end{equation}
where $J(\hat{\boldsymbol{x}}[t])$ is a convex cost function, $C()$ is the criterion or the error measure that the adaptive filter aims to minimize, usually being implemented by a norm calculation.
In \eqref{eq_optimization}, the condition $\mathbf{U\Sigma U}^{T}\hat{\boldsymbol{x}}[t] = \hat{\boldsymbol{x}}[t]$ is the bandlimited assumption \cite{bib_LMS}.
Since $J(\hat{\boldsymbol{x}}[t])$ is convex, we can obtain the results of \eqref{eq_optimization} using gradient descent, leading to a generalized graph adaptive filter update scheme
\begin{equation}  
 \hat{\boldsymbol{x}}[t+1] = \hat{\boldsymbol{x}}[t]+ \mu \frac{\partial J(\hat{\boldsymbol{x}}[t])} {\partial  \hat{\boldsymbol{x}}[t]},
\end{equation}
where $\mu$ is a parameter that controls the amount of the update $\frac{\partial J(\hat{\boldsymbol{x}}[t])} {\partial  \hat{\boldsymbol{x}}[t]}$ at each time instance. 

We can see that the core of this update lies in the effectiveness of the convolution $\mathbf{U\Sigma U}^{T}$.
For a time-invariant graph structure, the matrix $\mathbf{U}$ is defined by the graph structure, which leaves the only room of freedom to tune the convolution in the spectral filter $\mathbf{\Sigma}$.
Ideally, the predefined filter should have the same spectrum as the ground truth signal.
However, an effective filter pre-definition usually requires human prior knowledge to design.
This gives rise to a necessity for learning an effective filter from the noisy and missing graph signals.
In LMP-GNN, we intend to solve this problem using the learning capability of GCN.

\subsection{Least Mean $p^{th}$ Power Graph Neural Network}
Now that we have properly defined the graph adaptive filters and the GCNs, we will proceed to the derivations of the LMP-GNN.
Remember we would like the LMP-GNN to perform the online prediction of time-varying graph signals when the observation is missing and under noise pollution. 
The noise is possibly impulsive, so the LMP-GNN should be robust.
Thus, we first modify the cost function in \eqref{eq_optimization} to give a robust result under impulsive noise. 
The choice of $C()$ is to be the $l_p$-norm, as $l_p$-norm will output a result following the minimum dispersion criterion that is proven to be stable under impulsive noise \cite{nguyen2020_LMP, bib_MD_LMAD}:
\begin{equation}
\begin{aligned}
\min_{\hat{\boldsymbol{x}}[t]} \quad &  J(\hat{\boldsymbol{x}}[t]) =  \mathbb{E}\lvert\mathbf{D}_{\mathcal{S}}(\boldsymbol{y}[t]-\mathbf{U\Sigma U}^{T}\hat{\boldsymbol{x}}[t])\rvert ^p,\\
\textrm{s.t.} \quad & \mathbf{U\Sigma U}^{T}\hat{\boldsymbol{x}}[t] = \hat{\boldsymbol{x}}[t],\\
\end{aligned}
\label{eq_optimization_LMP}
\end{equation}
where $\mathbb{E}$ is the expectation and superscript $p$ is the element-wise $p^{th}$ power with $1\leq p \leq2$.
The results of \eqref{eq_optimization_LMP} is
\begin{equation}
    \mathbf{U}\sum_{f=1}^{F}\mathbf{\Theta}_{f, l}\mathbf{U}^T\lvert\boldsymbol{\epsilon}[t]\rvert^{p-1}\circ\text{Sign}(\boldsymbol{\epsilon}[t]),
    \label{eq_opt_result_lmp}
\end{equation}
with
\begin{equation}
    \boldsymbol{\epsilon}[t] = \mathbf{D}_{\mathcal{S}}(\boldsymbol{y}[t]-\hat{\boldsymbol{x}}[t])
    \label{eq_error}
\end{equation} 
being the estimation error of the observed signal, $\circ$ is the element-wise multiplication, and Sign$()$ is the sign function.

Now, adding the optimization results from \eqref{eq_opt_result_lmp} with the self-aggregation of the estimated signal $\hat{\boldsymbol{x}}[t]$, and feeding this sum into an activation function $\sigma$, we can form the forward propagation scheme of the LMP-GNN. 
A single layer of LMP-GNN is formulated as
\begin{equation}
        \hat{\boldsymbol{x}}_{l+1}[t] =\sigma\left(\hat{\boldsymbol{x}}_{l}[t]+\mu \mathbf{U}\sum_{f=1}^{F}\mathbf{\Theta}_{f, l}\mathbf{U}^T\lvert\boldsymbol{\epsilon}[t]\rvert^{p-1}\circ\text{Sign}(\boldsymbol{\epsilon}[t])+\boldsymbol{b}_l[t]\right).
    \label{LMP_GNN_layer}
\end{equation}
Following the formulation convention of GCN, we introduce a bias term $\boldsymbol{b}_l[t]$ within the activation function of the LMP-GNN.

The utilization of the minimum dispersion criterion in LMP-GNN allows it to robustly perform online estimations under different impulsive noise levels by giving different parameter settings of the parameter $p$ \cite{bib_MD_LMAD}. 
This makes the LMP-GNN a universal framework combining adaptive filter and graph neural networks.
LMP-GNN inherently possesses the attributes of graph adaptive filters, allowing it to update predictions based on the error between prediction and observation in real time.
Rather than using static predefined filters as in prior adaptive GSP algorithms such as the GLMS and GLMP, the LMP-GNN in \eqref{LMP_GNN_layer} inherits the capability of neural networks to train and learn filter parameters $\sum_{f=1}^{F}\mathbf{\Theta}_{f, l}$ directly from the graph signals.
In the backward propagation phase of the LMP-GNN, the gradient for updating the network weights $\sum_{f=1}^{F}\mathbf{\Theta}_{f, l}$ and the bias term $\boldsymbol{b}_l[t]$ is calculated using the error term $\boldsymbol{\epsilon}[t]$ in \eqref{eq_error}, along the path that $\hat{\boldsymbol{x}}_l[t]$ propagates. 
We should point out that because the LMP-GNN resembles the update strategy of the classical adaptive filters, the LMP-GNN has the capability of updating its parameters as it makes predictions. 
Additionally, the LMP-GNN can conduct online estimations as same as adaptive filters.
This is different than graph adaptive filters such as the GLMS algorithm, the GNLMS algorithm, the GLMP algorithm, the GNLMP algorithm, and the G-Sign algorithm, as they all rely on predefined graph filters and do not learn or change the filters during their operation.
Ideally, LMP-GNN is able to obtain a filter that is closer to the ground truth graph signal. 

\subsection{Special Cases of the LMP-GNN}
In classical adaptive filters, the adaptive LMP algorithm is a generalized formulation for the adaptive LMS algorithm and the adaptive Sign filter: setting $p = 2$ will lead to the LMS algorithm, and setting $p = 1$ will lead to the Sign algorithm \cite{bib_classical_adaptive_filter}.
Similarly, we can view the LMP-GNN as the generalized formulation for the Least Mean Squares Graph Neural Networks (LMS-GNN) and the Sign Graph Neural Networks (Sign-GNN).

The LMS-GNN has recently been proposed in \cite{yan_2024_LMS_GNN} as that is also aimed to solve the same problem as the LMP-GNN at the online estimation of time-varying graph signals. 
However, LMS-GNN is the generalization of LMP-GNN in two aspects. 
First, the $l_p$-norm optimization problem in \eqref{eq_optimization_LMP} and its results in \eqref{eq_opt_result_lmp} are the generalized form for the $l_2$-norm used in LMS-GNN. 
This gives the same LMP-LMS analogy found in classical adaptive filters. 
Secondly, the minimum dispersion criterion reduces to the minimum mean square error when $p = 2$ for Gaussian noise, but for $ 1\leq p < 2$ the minimum dispersion criterion is more suitable for other noise distributions other than the Gaussian distribution \cite{bib_MD_LMAD}. 
In fact, setting $p = 1$ can remove the assumption on noise distributions, as we will be discussing in the next part of this subsection.
For the details of the LMS-GNN algorithm, please refer to \cite{yan_2024_LMS_GNN} as the theme of this paper is on the more universal and more robust LMP-GNN. 

Notice that when $p=1$, our proposed LMP-GNN will become a special form which we call Sign-GNN.
To obtain the Sign-GNN, we turn our attention again to the optimization problem in \eqref{eq_optimization} but this time with $p=1$:
\begin{equation}
\begin{aligned}
\min_{\hat{\boldsymbol{x}}[t]} \quad &  J(\hat{\boldsymbol{x}}[t]) =  \mathbb{E}\lvert\mathbf{D}_{\mathcal{S}}(\boldsymbol{y}[t]-\mathbf{U\Sigma U}^{T}\hat{\boldsymbol{x}}[t]))\rvert ^1,\\
\textrm{s.t.} \quad & \mathbf{U\Sigma U}^{T}\hat{\boldsymbol{x}}[t] = \hat{\boldsymbol{x}}[t].\\
\end{aligned}
\label{eq_optimization_Sign}
\end{equation}
Solving \eqref{eq_optimization_Sign} will give us a solution 
\begin{equation}
    \mathbf{U}\sum_{f=1}^{F}\mathbf{\Theta}_{f, l}\mathbf{U}^T\text{Sign}(\boldsymbol{\epsilon}[t]).
    \label{eq_opt_result_sign}
\end{equation}
Following the previous derivations, the forward propagation scheme of one Sign-GNN layer is
\begin{equation}
        \hat{\boldsymbol{x}}_{l+1}[t] =\sigma\left(\hat{\boldsymbol{x}}_{l}[t]+\mu \mathbf{U}\sum_{f=1}^{F}\mathbf{\Theta}_{f, l}\mathbf{U}^T\text{Sign}(\boldsymbol{\epsilon}[t])+\boldsymbol{b}_l[t]\right).
    \label{Sign_GNN_layer}
\end{equation}
It should be noticed that the sign function only outputs three values, -1, 0, +1, and in reality, the elements within $\boldsymbol{\epsilon}[t]$ are rarely 0 as $\boldsymbol{\epsilon}[t]$ is the error term. 
As a result, the sign function essentially clips the error, making it insensitive to the estimation error caused by the noise, especially the outliers caused by the impulsive noise \cite{bib_sign_clip}.
Looking at another perspective, the Sign-GNN removes any noise assumptions.
The formulation in \eqref{Sign_GNN_layer} also reduces the computational complexity of the Sign-GNN when compared to LMP-GNN in \eqref{LMP_GNN_layer} because the calculation of $(p-1)^{th}$ power is removed from the forward propagation of the Sign-GNN in \eqref{Sign_GNN_layer}.
These two characteristics of the Sign-GNN make Sign-GNN stand out as a special case of the LMP-GNN.

\section{Experiments}
\label{sec_experiments}
We conduct adequate experiments on two time-varying graph signal datasets to test our proposed method and compare it with previous baselines. Other than S$\alpha$S noise, we test various noise of different distributions to prove the effectiveness and robustness of our proposed method.

\subsection{Experiment Setup}

The performance of the LMP-GNN at predicting time-varying graph signals will be tested under two real-world datasets: the temperature dataset \cite{bib_dataset} and the traffic speed dataset \cite{cui2019traffic}. 
The temperature dataset is a time-varying graph signal of $T = 95$ hourly temperatures collected from $N=197$ weather stations across the U.S. The traffic speed dataset is a time-varying graph signal of $T = 288$ recordings of vehicle speed collected from $N=157$ inductive loop detectors on freeways around the Seattle area, with the time interval of two recordings being 5 minutes. For the temperature dataset, the graph structure is formed according to the 7-nearest-neighbor and Gaussian Kernel approach based on the latitude and longitude of the weather stations seen in \cite{Spelta_2020_NLMS}. For the traffic dataset, the graph structure is provided from the source. We will be comparing LMP-GNN and its variant Sign-GNN with the following baseline algorithms: the GLMS algorithm \cite{bib_LMS}, the GLMP algorithm \cite{nguyen2020_LMP}, the GNLMS algorithm \cite{Spelta_2020_NLMS}, the GNLMP algorithm \cite{nguyen2020_LMP},  the G-Sign algorithm \cite{yan_2022_sign}, the LMS-GNN \cite{yan_2024_LMS_GNN}, the GCN \cite{kipf2016semi}, and the STGCN \cite{yu2018_STGCN}. The chosen baselines are comprehensive and compelling for containing both graph adaptive filters (GLMS, GLMP, GNLMS, GNLMP, G-Sign) and GNN-based (GCN, STGCN, LMS-GNN) methods.

For the GNN-based algorithms, we partition a portion of the whole dataset as the training set used to train and learn the neural network weights. For the temperature dataset, the training set contains the first 24 recordings. For the traffic dataset, the training set contains the first 72 recordings. In both cases, the training sets comprise approximately a quarter of the complete dataset. As for the graph adaptive filtering algorithms, we refer to \cite{Spelta_2020_NLMS} to define bandlimited filters for GLMP, GNLMP, GLMS, GNLMS, and G-Sign using the spectrum of the training set. The bandlimited filter definition follows a greedy approach that maximizes the spectral content to keep and eliminates the rest. In the temperature dataset, the number of frequency content to keep in the bandlimited filter is $|\mathcal{F}| = 120$; for the traffic dataset the kept frequency content is $|\mathcal{F}| = 80$. 

To better test the robustness of the LMP-GNN under various noise conditions, we selected four different noise distributions of various scales. The experiments are conducted in separate runs and are repeated $R = 100$ repetitions for each noise setting. For S$\alpha$S noise, we set $\gamma = 0.1$ and $\alpha=1.1,1.2,1.4,1.6,1.8,2.0$. For Cauchy noise, we set $\gamma=0.1$, which is a special case of S$\alpha$S noise. For Student's t noise, we set $\nu=10$. For Laplace noise, we set $b=3$. All the noises are zero-mean and regarded as additive noise to the ground truth graph signal. The observation ratio of the temperature dataset is $130/197$, while the observation ratio of the traffic dataset is $100/157$. The observed and missing graph signals the same spatial sampling strategy seen in \cite{Spelta_2020_NLMS}. Combining noise with the missingness of the data, the input observations to all the algorithms will follow the model in \eqref{eq_observation}.

\subsection{Results and Discussion}

\begin{figure}[h]
    \centering
        \begin{subfigure}{\linewidth}
         \centering
         \includegraphics[width=\textwidth]{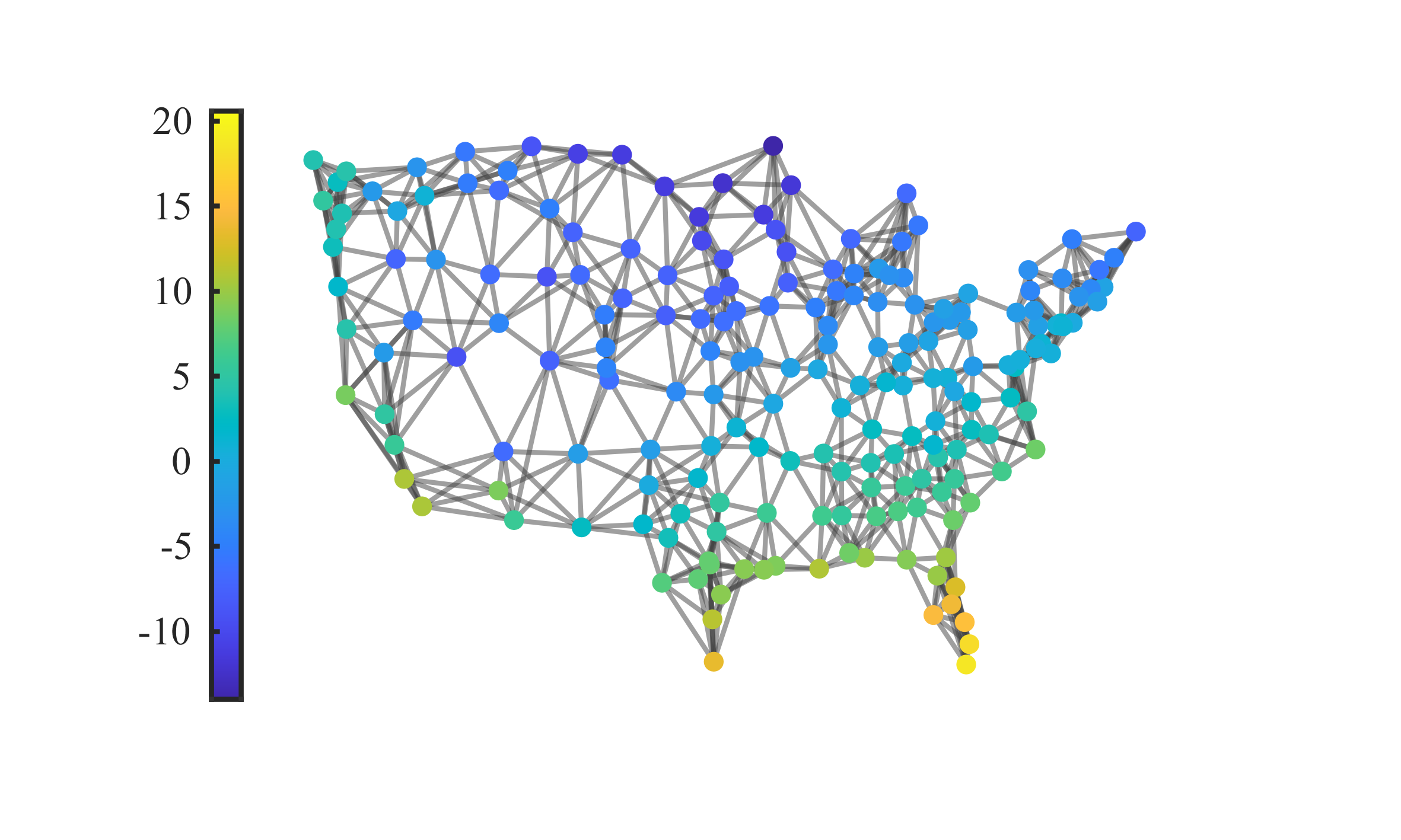}
        \end{subfigure}
             
     \vspace{-30 pt}

        \begin{subfigure}{\linewidth}
         \centering
         \includegraphics[width=\textwidth]{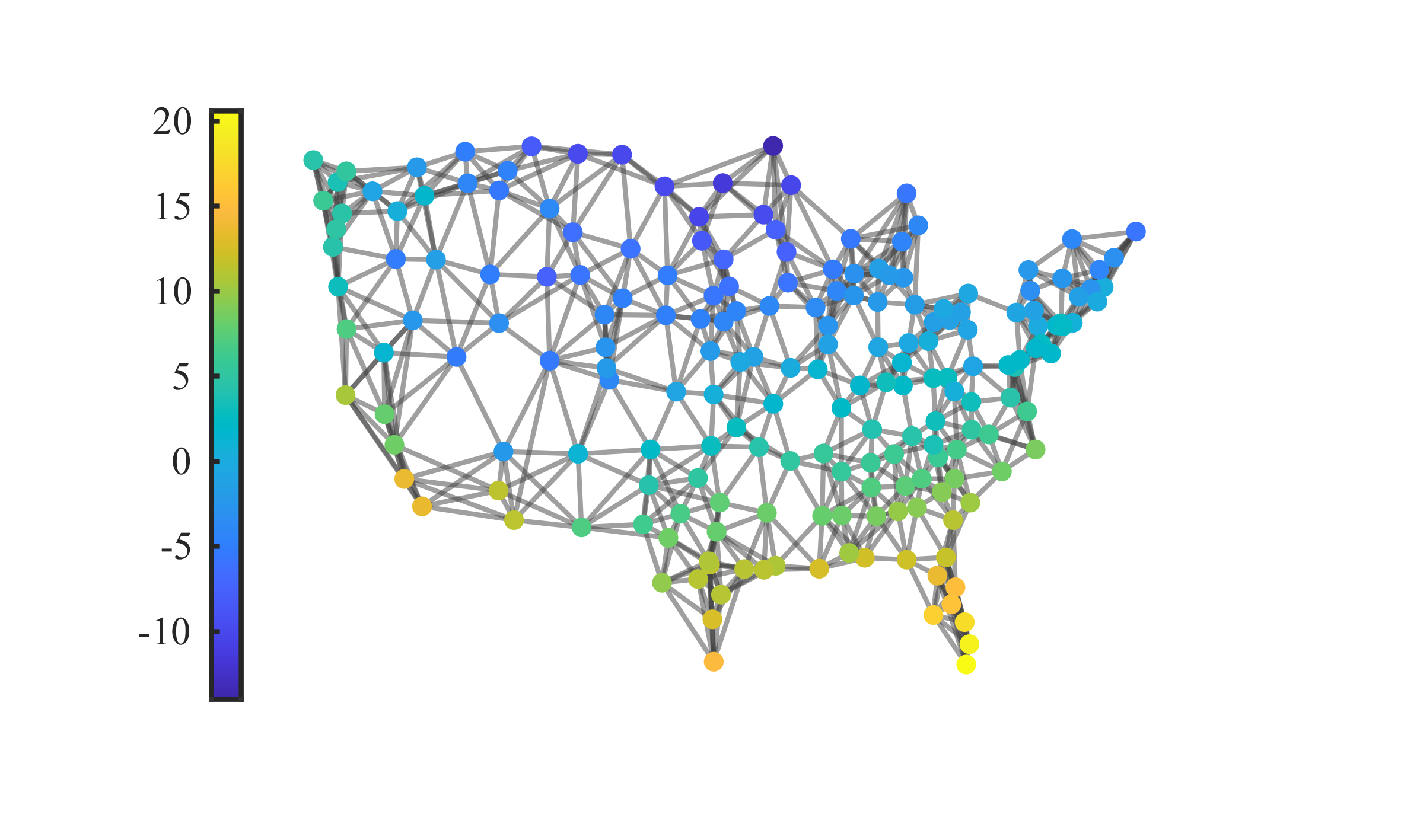}
        \end{subfigure}
             
     \vspace{-30 pt}
     
        \begin{subfigure}{\linewidth}
         \centering
         \includegraphics[width=\textwidth]{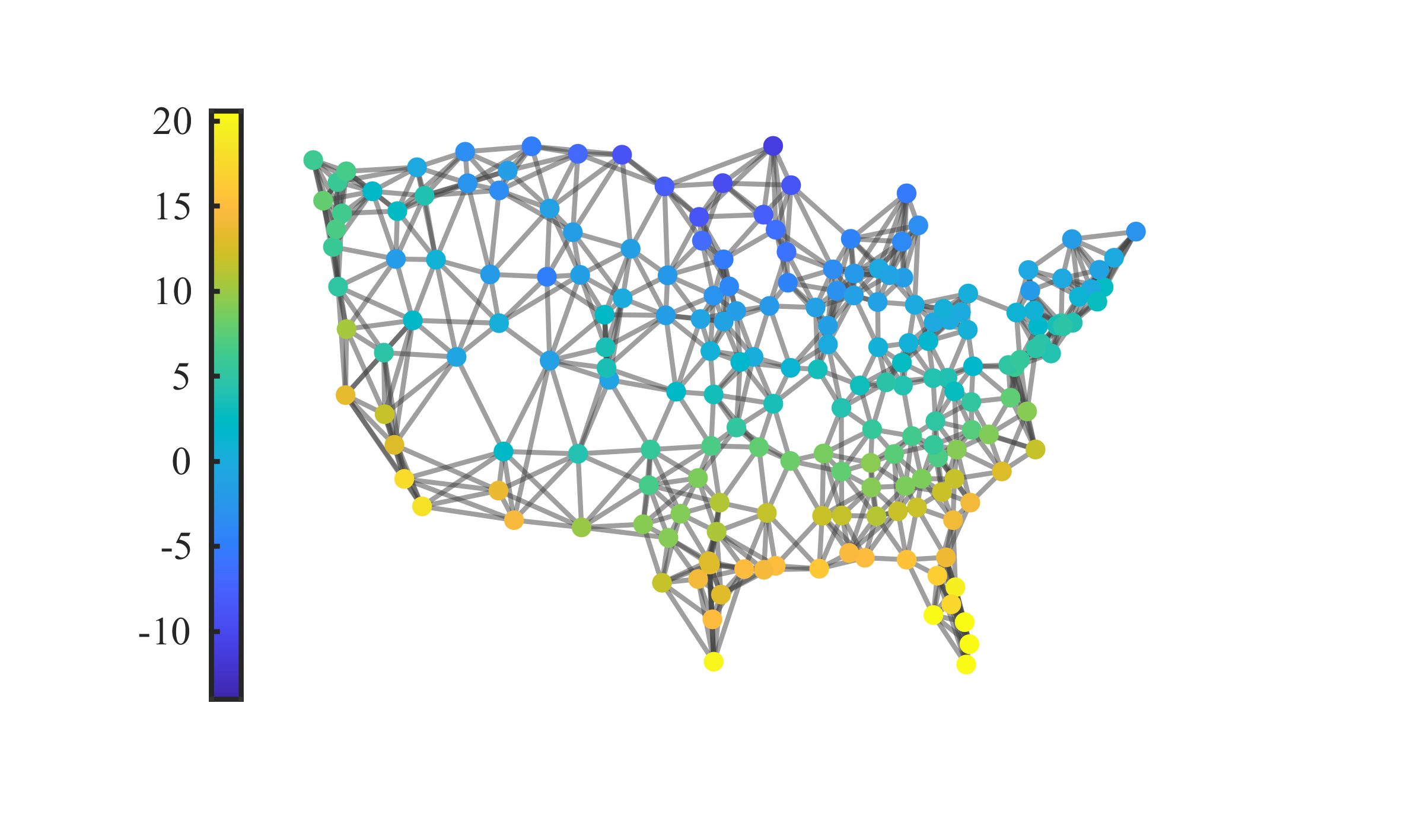}
        \end{subfigure}
             
     \vspace{-20 pt}
     
    \caption{Restored time-varying temperature recordings from the missing and noisy observations using the LMP-GNN. The three displayed time instances are the same as in Figure~\ref{fig_ground_truth}. }
    \label{fig_pred}
\end{figure}

We measure the performance of all methods by the Mean Squared Error (MSE) at each time step
\begin{equation}
    \text{MSE}[t] = \frac1N\sum_{i=1}^N\\{(x_i[t]-\hat{x}_i[t])^2},
\end{equation}
where $N$ denotes the number of nodes, $x_i[t]$ is the ground truth value at the $i^{th}$ node while $\hat{x}_i[t]$ is the predicted value at the $i^{th}$ node. The $\text{MSE}[t]$ of all methods under S$\alpha$S noise is shown in Table~\ref{table_MSE_temperature_alphastable} for the temperature dataset and Table~\ref{table_MSE_traffic_alphastable} for the traffic dataset. 
As for the Cauchy noise, Student's t noise, and Laplace noise, the MSE$[t]$ of the temperature dataset is recorded in Table~\ref{table_MSE_temperature_others} and the MSE$[t]$ of the traffic dataset are summarized in Table~\ref{table_MSE_traffic_others}.

\begin{table*}[h]
    \centering
    \caption{Prediction spatial MSE averaged over all the time points of temperature dataset under S$\alpha$S noise. Methods of best performance and the second best performance are denoted in bold and underlined fonts respectively}
    \begin{tabular}{c c c c c c c c c c c}
        \hline
         &LMP-GNN &LMS-GNN &Sign-GNN &GLMP &GNLMP &GLMS &GNLMS  &G-Sign &GCN &STGCN  \\
        \hline
        $\alpha$ = 1.0 & \multicolumn{10}{c}{(see Cauchy results)} \\
        \hline
        $\alpha$ = 1.1 &\bf{2.15} &3.88  &2.83  &20.94  &5.61  &4.68  &7.27  &\underline{2.80}  &9.18  &5.88  \\
        \hline
        $\alpha$ = 1.2 & \bf{1.92} &3.02  &2.80  &20.36  &4.43  &4.60  &7.20  &\underline{2.78}  &8.20  &5.60  \\
        \hline
        $\alpha$ = 1.4 & \underline{1.68} &\bf{1.40}  &2.88  &19.49  &2.34  &3.17  &5.59  &2.77  &8.06  &5.25  \\
        \hline
        $\alpha$ = 1.6 & \underline{1.50} &\bf{1.26}  &2.95  &18.98  &1.89  &3.12  &5.55  &2.77  &8.05  &5.86  \\
        \hline
        $\alpha$ = 1.8 & \underline{1.40} &\bf{1.26}  &3.02  &18.64  &1.73  &3.12  &5.54  &2.76  &8.06  &6.98  \\
        \hline
        $\alpha$ = 2.0 & \bf{1.30} &\underline{1.33}  &3.07  &18.41  &1.69  &3.11  &1.66  &2.77  &8.11  &5.55  \\
        \hline
    \end{tabular}
    \label{table_MSE_temperature_alphastable}
\end{table*}

\begin{table*}[h]
    \centering
    \caption{Prediction spatial MSE averaged over all the time points of temperature dataset under various noise. Methods of best performance and the second best performance are denoted in bold and underlined fonts respectively}
    \begin{tabular}{c c c c c c c c c c c}
        \hline
         &LMP-GNN &LMS-GNN &Sign-GNN &GLMP &GNLMP &GLMS &GNLMS  &G-Sign &GCN &STGCN  \\
        \hline
        Cauchy &\bf{2.47} &8.68  &3.15  &20.97  &8.92  &9.22  &11.80  
        &\underline{2.81}  &20.28  &11.20  \\
        \hline
        Student's t &\bf{1.30}  &\underline{1.39}  &1.78  &21.13  &2.98  &3.12  &5.55 &3.01  &8.06 &5.30  \\
        \hline
        Laplace &\bf{1.88}  &2.70 &\underline{2.17}  &21.72  &4.73  &3.19  &5.59  &4.51  &8.10  &6.47  \\
        \hline
    \end{tabular}
    \label{table_MSE_temperature_others}
\end{table*}

\begin{table*}[h]
    \centering
    \caption{Prediction spatial MSE averaged over all the time points of traffic dataset under S$\alpha$S noise. Methods of best performance and the second best performance are denoted in bold and underlined fonts respectively}
    \begin{tabular}{c c c c c c c c c c c}
        \hline
         &LMP-GNN &LMS-GNN &Sign-GNN &GLMP &GNLMP &GLMS &GNLMS  &G-Sign &GCN &STGCN  \\
        \hline
        $\alpha$ = 1.0 & \multicolumn{10}{c}{(see Cauchy results)} \\
        \hline
        $\alpha$ = 1.1 &8.73 &12.70  &\bf{6.52}  &27.67  &\underline{7.76}  &17.27  &10.27  &8.39  &209.33  &67.10  \\
        \hline
        $\alpha$ = 1.2 & \underline{7.68} &9.79  &\bf{5.95}  &28.34  &7.72  &9.86  &8.14  &8.42  &184.77  &71.43  \\
        \hline
        $\alpha$ = 1.4 & \underline{7.37} &8.81  &\bf{5.76}  &31.04  &7.68  &10.86  &8.28  &8.47  &180.85  &41.39  \\
        \hline
        $\alpha$ = 1.6 & \underline{7.44} &8.71  &\bf{5.78}  &35.96  &7.66  &8.57  &7.89  &8.51  &173.57  &34.38  \\
        \hline
        $\alpha$ = 1.8 & \underline{7.66} &9.37  &\bf{5.82}  &44.46  &7.69  &8.47  &7.88  &8.53  &178.78  &24.60  \\
        \hline
        $\alpha$ = 2.0 & 8.58 &8.88  &\bf{5.80}  &59.73  &\underline{7.79}  &8.47  &7.87  &8.55  &179.75  &41.49  \\
        \hline
    \end{tabular}
    \label{table_MSE_traffic_alphastable}
\end{table*}

\begin{table*}[h]
    \centering
    \caption{Prediction spatial MSE averaged over all the time points of traffic dataset under various noise. Methods of best performance and the second best performance are denoted in bold and underlined fonts respectively}
    \begin{tabular}{c c c c c c c c c c c}
        \hline
         &LMP-GNN &LMS-GNN &Sign-GNN &GLMP &GNLMP &GLMS &GNLMS  &G-Sign &GCN &STGCN  \\
        \hline
        \hline
        Cauchy &  12.60 &32.88  &9.84  &27.72  &\bf{7.82}  &48.98  &18.81  &\underline{8.34}  &302.26  &82.41  \\
        \hline
        Student's t & \underline{7.46} &8.14  &\bf{5.80}  &39.30  &7.61  &8.64  &7.94  &8.08  &160.69  &45.45  \\
        \hline
        Laplace & 10.12 &11.72  &\bf{6.67}  &70.85  &\underline{7.72}  &8.98  &8.02  &\underline{7.72}  &146.36  &46.74  \\
        \hline
    \end{tabular}
    \label{table_MSE_traffic_others}
\end{table*}

\begin{figure}[h]
    \centering
    \includegraphics[width=\linewidth, trim={0 0 0 0},clip]{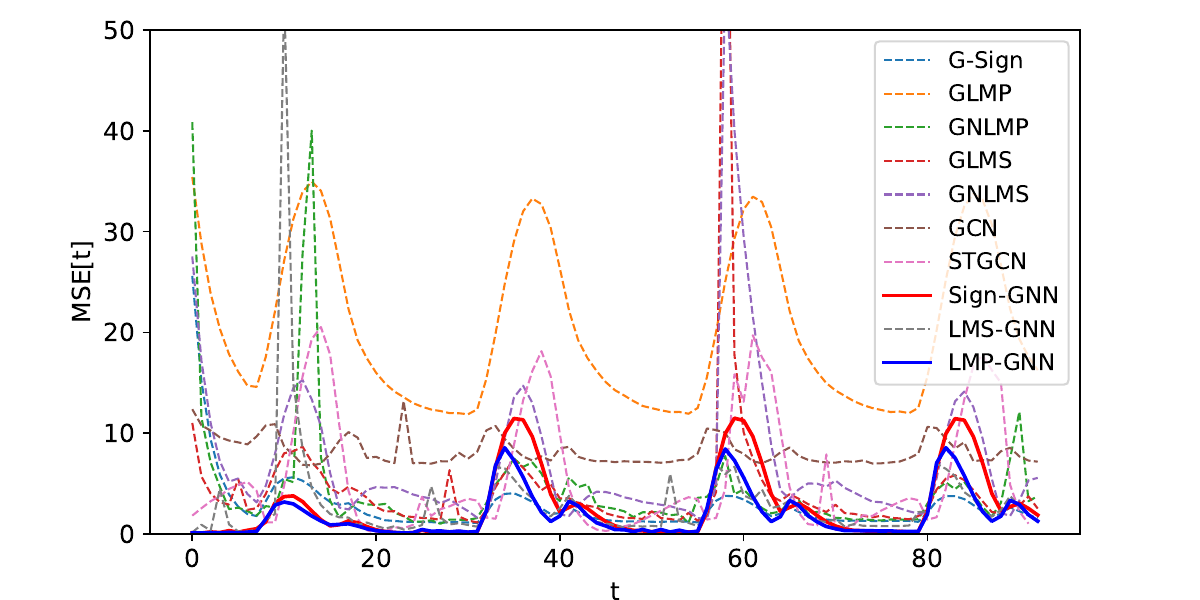}
    \caption{The MSE at each time instance of the temperature dataset under S$\alpha$S noise with $\alpha = 1.2$ and $\mu = 0.1$.}    
    \label{fig_mse_temperature}
\end{figure}

\begin{figure}[h]
    \centering
    \includegraphics[width=\linewidth, trim={0 0 0 0},clip]{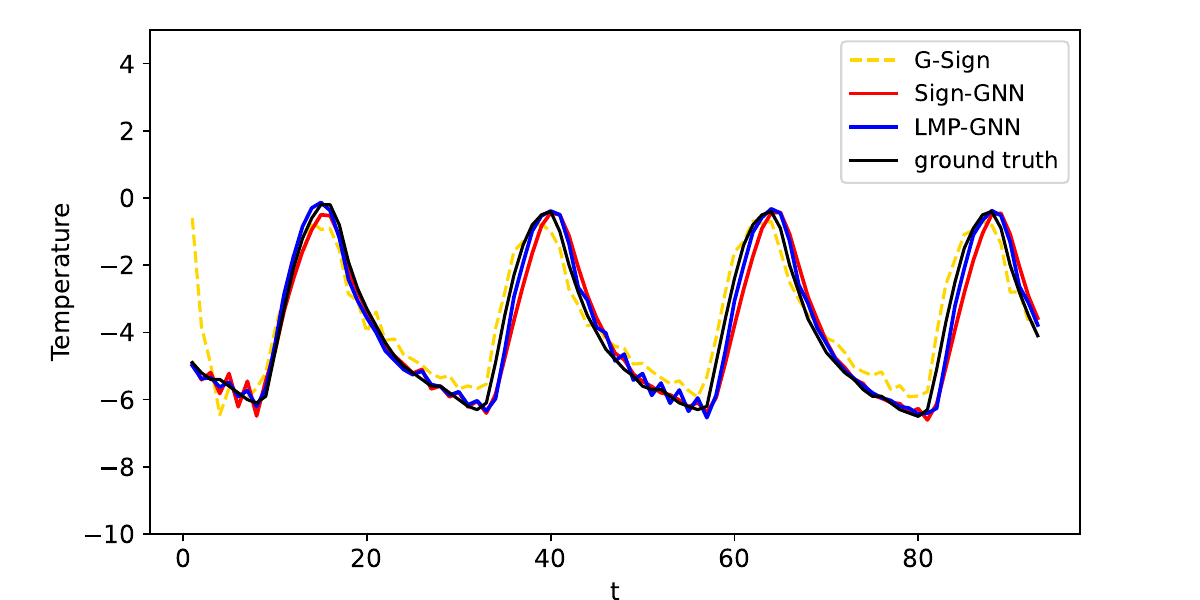}
    \caption{A showcase of the prediction results at one single node of the temperature dataset.}    
    \label{fig_prediction_temperature}
\end{figure}

In order to visually highlight the performance of all methods at each time instance, we plot the $\text{MSE}[t]$ of all methods for the S$\alpha$S noise when $\alpha = 1.2$ and $\mu = 0.1$ in Figure ~\ref{fig_mse_temperature}. To provide a more intuitive representation of the prediction results, we also give an example of the prediction results of the three best methods under S$\alpha$S noise when $\alpha=1.2$ of the temperature dataset in Figure ~\ref{fig_prediction_temperature}. The prediction results of the entire temperature datasets at 3 different time instances done by the LMP-GNN are also shown in Figure~\ref{fig_pred} to provide a more straightforward visualization.

For the temperature dataset, from Table~\ref{table_MSE_temperature_alphastable} and Table~\ref{table_MSE_temperature_others}, we can see that our LMP-GNN has either the lowest MSE or the second lowest MSE under all value settings of $\alpha$, and other noise as well. 
When comparing the LMP-GNN with graph adaptive filters, namely the GLMS algorithm, the GLMP algorithm, the GNLMS algorithm, the GNLMP algorithm, and the G-sign algorithm, we see that the LMP-GNN outperforms all the graph adaptive filters.
The good performance of the LMP-GNN is attributed to the learning power of the LMP-GNN: instead of predefined a filter, the LMS-GNN learns a more effective filter than the predefined one. 
LMS-GNN also performs well especially when facing S$\alpha$S noise with a higher value of $\alpha$, which corresponds to less impulsiveness scenarios. 
We should point out that the LMS-GNN is a special case of the LMP-GNN, so the LMS-GNN outputs can always be obtained from the LMP-GNN by setting $p = 2$.
However, when $\alpha$ has a lower value that leads to high impulsiveness behavior of the S$\alpha$S noise, the LMS-GNN exhibits a significant performance degradation due to its instability under impulsive noise.
The LMP-GNN maintains a relatively favorable and stable performance. 
This is due to the design of LMP-GNN, the $l_p$-norm optimization is designed with the consideration of robustness under impulsive noise, while LMS-GNN focuses on $l_2$-norm optimization that has the Gaussian (non-impulsive) assumption. Such difference results in LMP-GNN's robustness against noise of high variance. 
It is worth noting that GCN and STGCN perform worse than most adaptive filters, namely the G-Sign algorithm and the GNLMP algorithm. 
For GCN, the reason is that it considers only graph relations but does not take into account the time-varying relations. 
For STGCN, the careful design of network architecture indeed allows it to better capture spatio-temporal relationships, but it also has a higher requirement for training data. 
To properly train STGCN, a huge amount of training data from a longer time period should be fed for better performance. 
However, in our experiment setting the training data is limited, noisy, and contains missingness. 
Besides, all methods are required to perform online estimation based on current one-step observation, which further constrains STGCN.
As a result, STGCN indeed performs better than GCN for considering spatio-temporal relations, but it fails to make better one-step predictions than adaptive filter-based methods. 
From Figure ~\ref{fig_mse_temperature}, we can find our proposed method LMP-GNN has the lowest MSE at almost every time step, which means LMP-GNN predicts more accurately than other methods at almost every time step, which proves the stability. 
Figure ~\ref{fig_prediction_temperature} shows the prediction results on one of the graph nodes of the three top-performing methods: LMP-GNN, Sign-GNN, and G-Sign, in which we can visually confirm the effectiveness of LMP-GNN.

As for the traffic dataset, the overall trend remains unchanged. Sign-GNN, as the variant of LMP-GNN, performs best under almost all noise settings. It is noteworthy that the MSE on the traffic dataset is generally higher than that on the temperature dataset. This could be attributed to the fact that the traffic dataset is collected from a circular freeway, resulting in a sparser graph structure, and consequently, the adjacency matrix of the graph may lack sufficient information between nodes.

In summary, LMP-GNN and its variant Sign-GNN demonstrate superior performance in online graph signal estimation under noisy and missing observations across datasets from different domains. The ability to capture space-time changes and handle various noises of different distributions makes it suitable for such real-world tasks.

\section{Conclusion}
\label{sec_conclusion}
We propose a universal framework LMP-GNN for online graph estimation under noisy and missing observations. Based on the adaptive filter backbone, LMP-GNN leverages graph neural network to online update filter weights, instead of previous predefined filters. It not only keeps the ability of adaptive filters to denoise and deal with missing observations but also incorporates the strength of graph neural network to train and learn automatically, thus leading to better capture of spatio-temporal relations of graph signals. The $ l_p$-norm-based update scheme grants LMP-GNN robustness against various impulsive noises of different distributions. Experiments on real-world temperature data and traffic data prove our methods are better than adaptive filters or GCNs.
\section*{Acknowledgements}
This work was supported by the Tsinghua Shenzhen International Graduate School Startup Fund under Grant QD2022024C and the Shenzhen Science and Technology Innovation Commission under Grant JCYJ20220530143002005. 
\bibliography{references}

\begin{thebibliography}{10}
\expandafter\ifx\csname url\endcsname\relax
  \def\url#1{\texttt{#1}}\fi
\expandafter\ifx\csname urlprefix\endcsname\relax\def\urlprefix{URL }\fi
\expandafter\ifx\csname href\endcsname\relax
  \def\href#1#2{#2} \def\path#1{#1}\fi

\bibitem{Ortega_2018}
A.~Ortega, P.~Frossard, J.~Kovačević, J.~M.~F. Moura, P.~Vandergheynst, Graph signal processing: Overview, challenges, and applications, Proceedings of the IEEE 106~(5) (2018) 808--828.

\bibitem{dong2020graph}
X.~Dong, D.~Thanou, L.~Toni, M.~Bronstein, P.~Frossard, Graph signal processing for machine learning: A review and new perspectives, IEEE Signal processing magazine 37~(6) (2020) 117--127.

\bibitem{Spelta_2020_NLMS}
M.~J.~M. Spelta, W.~A. Martins, Normalized lms algorithm and data-selective strategies for adaptive graph signal estimation, Signal Processing 167 (2020) 107326.

\bibitem{yu2018_STGCN}
B.~Yu, H.~Yin, Z.~Zhu, Spatio-temporal graph convolutional networks: A deep learning framework for traffic forecasting, in: International Joint Conference on Artificial Intelligence, 2018.

\bibitem{bib_LMS}
P.~Di~Lorenzo, S.~Barbarossa, P.~Banelli, S.~Sardellitti, Adaptive least mean squares estimation of graph signals, IEEE Transactions on Signal and Information Processing over Networks 2~(4) (2016) 555 -- 568.

\bibitem{brain_modeling}
W.~Huang, T.~A.~W. Bolton, J.~D. Medaglia, D.~S. Bassett, A.~Ribeiro, D.~Van De~Ville, A graph signal processing perspective on functional brain imaging, Proceedings of the IEEE 106~(5) (2018) 868 -- 885.

\bibitem{yan_2023_diffusion}
Y.~Yan, E.~E. Kuruoglu, Fast and robust wind speed prediction under impulsive noise via adaptive graph-sign diffusion, in: IEEE Conference on Artificial Intelligence, 2023, pp. 302--305.

\bibitem{zhao2023sequential}
F.~Zhao, E.~E. Kuruoglu, Sequential monte carlo graph convolutional network for dynamic brain connectivity, arXiv.

\bibitem{Mei_GVAR_2017}
J.~Mei, J.~M.~F. Moura, Signal processing on graphs: Causal modeling of unstructured data, IEEE Transactions on Signal Processing 65~(8) (2017) 2077--2092.

\bibitem{Isufi_GARMA_2017}
E.~Isufi, A.~Loukas, A.~Simonetto, G.~Leus, Autoregressive moving average graph filtering, IEEE Transactions on Signal Processing 65~(2) (2017) 274--288.

\bibitem{Hong_GGARCH_2023}
J.~Hong, Y.~Yan, E.~E. Kuruoglu, W.~K. Chan, Multivariate time series forecasting with garch models on graphs, IEEE Transactions on Signal and Information Processing over Networks 9 (2023) 557--568.

\bibitem{bib_classical_adaptive_filter}
P.~Diniz, Adaptive Filtering: Algorithms and Practical Implementation, Springer, 2008.

\bibitem{nguyen2020_LMP}
N.~Nguyen, K.~Do{\u{g}}an{\c{c}}ay, W.~Wang, Adaptive estimation and sparse sampling for graph signals in alpha-stable noise, Digital Signal Processing 105 (2020) 102782.

\bibitem{yan_2022_sign}
Y.~Yan, E.~E. Kuruoglu, M.~A. Altinkaya, Adaptive sign algorithm for graph signal processing, Signal Processing 200.

\bibitem{bib_GCN}
T.~N. Kipf, M.~Welling, Semi-supervised classification with graph convolutional networks, International Conference on Learning Representations.

\bibitem{bruna2013_spectral_GCN}
J.~Bruna, W.~Zaremba, A.~Szlam, Y.~LeCun, Spectral networks and locally connected networks on graphs, International Conference on Learning Representations.

\bibitem{song2020spatial}
C.~Song, Y.~Lin, S.~Guo, H.~Wan, Spatial-temporal synchronous graph convolutional networks: A new framework for spatial-temporal network data forecasting, in: Proceedings of the AAAI conference on artificial intelligence, Vol.~34, 2020, pp. 914--921.

\bibitem{li2021spatial}
M.~Li, Z.~Zhu, Spatial-temporal fusion graph neural networks for traffic flow forecasting, in: Proceedings of the AAAI conference on artificial intelligence, Vol.~35, 2021, pp. 4189--4196.

\bibitem{bib_underwater}
S.~Banerjee, M.~Agrawal, Underwater acoustic communication in the presence of heavy-tailed impulsive noise with bi-parameter {Cauchy-Gaussian} mixture model, Symposium on Ocean Technology (2013) 1--7.

\bibitem{bib_student_t_noise}
Q.~Li, Y.~Ben, S.~M. Naqvi, J.~A. Neasham, J.~A. Chambers, Robust student’s t based cooperative navigation for autonomous underwater vehicles, IEEE Transactions on Instrumentation and Measurement 67~(8) (2018) 1762--1777.

\bibitem{b17_PLC}
O.~Karakuş, E.~Kuruoglu, M.~Altinkaya, Modelling impulsive noise in indoor powerline communication systems, Signal, Image and Video Processing 14 (2020) 1655–1661.

\bibitem{idan2010cauchy}
M.~Idan, J.~L. Speyer, Cauchy estimation for linear scalar systems, IEEE transactions on automatic control 55~(6) (2010) 1329--1342.

\bibitem{herranz2004alpha}
D.~Herranz, E.~Kuruo{\u{g}}lu, L.~Toffolatti, An alpha-stable approach to the study of the p (d) distribution of unresolved point sources in cmb sky maps, Astronomy \& Astrophysics 424~(3) (2004) 1081--1096.

\bibitem{bib_MD_LMAD}
M.~Shao, C.~Nikias, Signal processing with fractional lower order moments: stable processes and their applications, Proceedings of the IEEE 81~(7) (1993) 986--1010.

\bibitem{kuruoglu1997new}
E.~Kuruoglu, C.~Molina, S.~Godsill, W.~Fitzgerald, A new analytic representation for the symmetric alpha-stable probability density function, in: Proceedings of the 5th World Meeting of the International Society for Bayesian Analysis (ISBA). ASA: American Statistical Association, 1997, pp. 229--233.

\bibitem{bib_sampling}
P.~Di~Lorenzo, P.~Banelli, E.~Isufi, S.~Barbarossa, G.~Leus, Adaptive graph signal processing: Algorithms and optimal sampling strategies, IEEE Transactions on Signal Processing 66~(13) (2018) 3584--3598.

\bibitem{kipf2016semi}
T.~Kipf, M.~Welling, Semi-supervised classification with graph convolutional networks, International Conference on Learning Representations.

\bibitem{defferrard2016_cheb}
M.~Defferrard, X.~Bresson, P.~Vandergheynst, Convolutional neural networks on graphs with fast localized spectral filtering, Advances in neural information processing systems 29.

\bibitem{yan_2024_LMS_GNN}
Y.~Yan, C.~Peng, E.~E. Kuruoglu, Adaptive least mean squares graph neural networks and online graph signal estimation (2024).
\newblock \href {http://arxiv.org/abs/2401.15304} {\path{arXiv:2401.15304}}.

\bibitem{bib_sign_clip}
V.~Bhatia, B.~Mulgrew, A.~T. Georgiadis, Stochastic gradient algorithms for equalisation in alpha-stable noise, Signal Processing 86~(4) (2006) 835--845.

\bibitem{bib_dataset}
M.~Palecki, I.~Durre, S.~Applequist, A.~Arguez, J.~Lawrimore, {U.S.} climate normals 2020: {U.S.} hourly climate normals (1991-2020), NOAA National Centers for Environmental Information.

\bibitem{cui2019traffic}
Z.~Cui, K.~Henrickson, R.~Ke, Y.~Wang, Traffic graph convolutional recurrent neural network: A deep learning framework for network-scale traffic learning and forecasting, IEEE Transactions on Intelligent Transportation Systems.

\end{thebibliography}

\end{document}